\documentclass[10pt,twocolumn,letterpaper]{article}

\usepackage{cvpr}              % To produce the CAMERA-READY version
\usepackage[dvipsnames,table,xcdraw]{xcolor}

\definecolor{cvprblue}{rgb}{0.21,0.49,0.74}
\usepackage[pagebackref,breaklinks,colorlinks,citecolor=cvprblue]{hyperref}

\usepackage{url}            % simple URL typesetting
\usepackage{booktabs}       % professional-quality tables
\usepackage{amsfonts}       % blackboard math symbols
\usepackage{nicefrac}       % compact symbols for 1/2, etc.
\usepackage{microtype}      % microtypography
\usepackage{amsmath}
\usepackage{multirow}
\usepackage{bm}
\usepackage{paralist} 
\usepackage{graphicx}
\usepackage{wrapfig}
\usepackage{bbding}
\usepackage{color}
\usepackage{cite}
\usepackage{amsthm,amsmath,amssymb}
\usepackage{mathrsfs}
\usepackage{stfloats}
\usepackage{amssymb}
\usepackage{adjustbox}
\renewcommand{\thefootnote}{\arabic{footnote}}
\usepackage[percent]{overpic}
\usepackage{enumitem}
\usepackage{relsize}
\usepackage{xcolor}
\usepackage{booktabs}
\usepackage{subfiles}

\usepackage[accsupp]{axessibility}
\def\ie{i.e.} % that is, in other words

\def\bes#1{\textbf{\color{black}\underline{#1}}}
\def\cite{\citep}

\definecolor{lightblue}{RGB}{173,216,230}
\definecolor{mydarkred}{RGB}{139,0,0}
\definecolor{c1}{HTML}{d6ecf0}

\def\paperID{6395} % *** Enter the Paper ID here
\def\confName{CVPR}
\def\confYear{2024}

\begin{document}

%%%%%%%%% TITLE - PLEASE UPDATE
%\title{Joint-learning Multi-Modal Salient and Camouflaged Object Detection}
%\title{General Multimodal Salient and Camouflaged Object Detection \\with 2D Prompt Learning}
\title{VSCode: General Visual Salient and Camouflaged Object Detection \\with 2D Prompt Learning}
\author{
	Ziyang Luo$^{1}$
	\hspace{5pt}
	Nian Liu$^{2,*}$
	\hspace{5pt}
	Wangbo Zhao$^{3}$
	\hspace{5pt}
	Xuguang Yang$^{1}$
	\hspace{5pt}
	Dingwen Zhang$^{1}$
	\hspace{5pt}\\
    Deng-Ping Fan$^{5,6}$
	\hspace{5pt}
    Fahad Khan$^{2,4}$
	\hspace{5pt}
    Junwei Han$^{1,7,*}$
	\\
	$^1$Northwestern Polytechnical University
	\hspace{6pt}
	$^2$Mohamed bin Zayed University of Artificial Intelligence
    \hspace{8pt}\\
    $^3$National University of Singapore
	\hspace{6pt}
    $^4$ CVL, Linköping University
	\hspace{6pt}
    $^5$ TBI Center, CS, Nankai University
	\hspace{6pt} \\
    $^6$ Nankai International Advanced Research Institute (SHENZHEN FUTIAN) 
        \hspace{8pt} \\
    $^7$ Institute of Artificial Intelligence, Hefei Comprehensive National Science Center 
}

\maketitle
\thispagestyle{empty}
\renewcommand{\thefootnote}{\fnsymbol{footnote}}
\footnotetext[1]{Corresponding author: Nian Liu (liunian228@gmail.com) and Junwei Han (junweihan2010@gmail.com)}

%第二版：从联合训练的角度出发，从共性和不同两个部分分开描述文章
\begin{abstract}
%%DP Fan
Salient object detection (SOD) and camouflaged object detection (COD) are related yet distinct binary mapping tasks. These tasks involve multiple modalities, sharing commonalities and unique cues. Existing research often employs intricate task-specific specialist models, potentially leading to redundancy and suboptimal results. We introduce VSCode, a generalist model with novel 2D prompt learning, to jointly address four SOD tasks and three COD tasks. We utilize VST as the foundation model and introduce 2D prompts within the encoder-decoder architecture to learn domain and task-specific knowledge on two separate dimensions. A prompt discrimination loss helps disentangle peculiarities to benefit model optimization. VSCode outperforms state-of-the-art methods across six tasks on 26 datasets and exhibits zero-shot generalization to unseen tasks by combining 2D prompts, such as RGB-D COD. Source code has been available at \href{https://github.com/Sssssuperior/VSCode}{https://github.com/Sssssuperior/VSCode}.
\end{abstract}

\vspace{-1mm}
\section{Introduction}
\label{sec:intro}
Visual salient object detection (SOD) and camouflaged object detection (COD) are two interconnected yet unique tasks. The goal of SOD is to identify prominent objects within an image that significantly contrast with their surroundings \cite{borji2019salient}, which can be used to promote segmentation \cite{zhao2023learning, li2023boosting}, detection \cite{xie2020count}, and Part-Object Relational visual saliency \cite{liu2022disentangled,liu2021part}. While COD focuses on identifying objects concealed within their environment. These objects intentionally blend in by sharing structural or textural similarities with their surroundings \cite{fan2020camouflaged}. Despite the seemingly different definitions of SOD and COD, they both belong to the realm of binary segmentation and share some vital fundamental similarities, such as objectness and structuredness.

\begin{figure}[t!]
    \centering
    \captionsetup{skip=5pt}
    \includegraphics[width=1\linewidth]{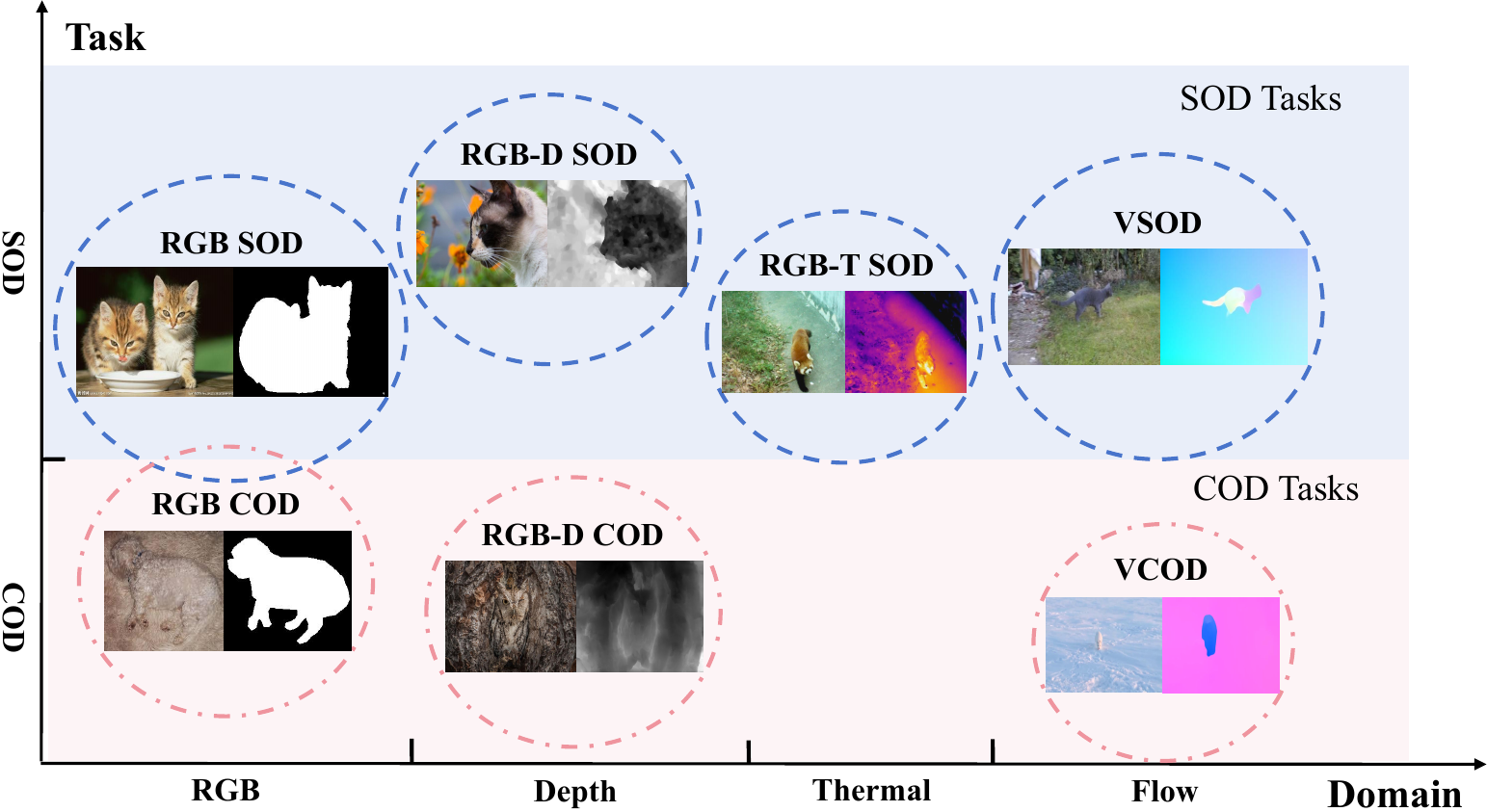}
    \caption{\textbf{Relationship of SOD, COD, and multimodal tasks.} Each specific task is seen as a combination of two dimensions, \ie\ domain (RGB/Depth/Thermal/Flow) and task (SOD/COD).}
    \label{relationTASK}
    \vspace{-0.5cm}
\end{figure}

To cater to various scenarios, both SOD and COD have given rise to several sub-tasks with different modalities, including RGB SOD \cite{zhang2018pagr,wang2017stagewise}, RGB COD \cite{ji2023deep,zhai2021mutual, pang2022zoom}, RGB-D SOD \cite{li2020icnet, Piao2019dmra}, RGB-D COD \cite{wu2023source}, and RGB-T SOD \cite{wang2021cgfnet, zhang2019rgb}.
By leveraging optical flow maps, Video SOD (VSOD) \cite{li2018flow,wei2020end} and VCOD \cite{lamdouar2020betrayed,cheng2022implicit} tasks can also be seen as a combination of two modalities. 
The relationship of SOD, COD, and multimodal tasks is shown in Figure~\ref{relationTASK}, where each specific task can be considered as a combination of two dimensions, \ie\, domain and task.
Although these multimodal tasks differ in the complementary cues they employ, these modalities share some key commonalities. For instance, depth, thermal, and optical flow maps often show obvious objectness as in RGB images.

Although previous CNN-based \cite{Piao2019dmra,zhao2019contrast,tu2022rgbt,wang2017video,lv2021simultaneously,cheng2022implicit} and transformer-based \cite{Liu_2021_ICCV,zhuge2021salient} approaches have effectively addressed these tasks and achieved favorable results, they usually rely on meticulously designed models to tackle each task individually. Designing models specifically for individual tasks can be problematic since the training data of one task is typically limited. Task-specific specialist models may be overly adapted to a particular task and overfitted to specific training data distribution, which ultimately sacrifices generalization ability and results in suboptimal performance. 
One solution may be using more data, however, being costly and time-consuming for data annotation. 
To this end, joint learning a generalist model emerges as a more promising option, as it allows for the maximum use of all data and the effective learning of the commonalities of all tasks, hence significantly reducing the risk of overfitting and enhancing the generalization capability \cite{kendall2018multi,liu2019end}.
However, joint learning multiple tasks is not straightforward. On one hand, simultaneously handling both commonalities and peculiarities of all tasks poses a significant challenge as the incompatibility among different tasks easily leads to a decline in performance with simple joint training \cite{li2023joint}.
On the other hand, it usually introduces additional complexity, computational costs, and parameters.

\begin{figure}[!t]
    \centering
    \captionsetup{skip=5pt}
    \includegraphics[width=1\linewidth]{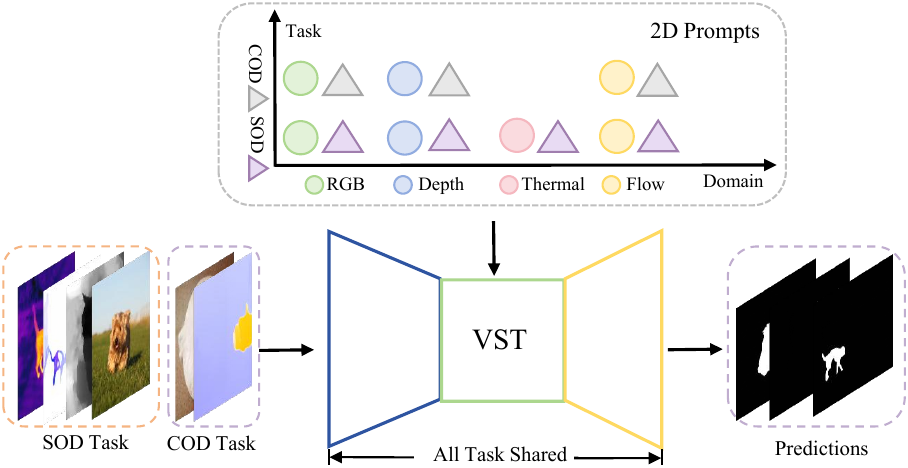}
    \caption{\textbf{Overall architecture of our VSCode model.} We use VST \cite{Liu_2021_ICCV} as the foundation model to acquire commonalities among multimodal SOD and COD tasks. For each task, we integrate 2D prompts to aggregate peculiarities along the domain dimension and the task dimension, including four domain-specific prompts and two task-specific prompts. 
    }
    \label{network}
    \vspace{-4mm}
\end{figure}

In this paper, we present a general \textbf{V}isual \textbf{S}alient and \textbf{C}amouflaged \textbf{o}bject \textbf{de}tection (VSCode) model which encapsulates both commonalities and peculiarities of different tasks with a simple but effective design, as illustrated in Figure~\ref{network}. 
On one hand, we adopt VST \cite{Liu_2021_ICCV} as the shared segmentation foundation model to assimilate commonalities of different tasks by leveraging its simple and pure-transformer-based architecture. On the other hand, inspired by the recent emergence of the parameter-efficient prompting technique \cite{nie2022pro, jia2022visual,zhao2024dynamic}, we propose 2D prompts to capture task peculiarities. Specifically, we decompose these peculiarities along the domain dimension and the task dimension, and consequently design domain-specific prompts and task-specific prompts to comprehend the differences among diverse domains and tasks, respectively. These 2D prompts can effectively disentangle domain and task peculiarities, making our model easily adaptable by combining them to tackle specific tasks and even unseen ones. Furthermore, we present a prompt discrimination loss to encourage the 2D prompts to focus on acquiring adequate peculiarities and enable the foundational model to concentrate on commonality learning. 

Finally, we train our VSCode model on four SOD tasks and two COD tasks, demonstrating its effectiveness against state-of-the-art methods. What's more, we carry out evaluations on a reserved task and reveal remarkable zero-shot generalization ability of our model, which has never been explored in previous works.
The main contributions in this work can be summarized as follows:
\begin{compactitem}
\item We present VSCode, the first generalist model for multimodal SOD and COD tasks.
\item We propose to use a foundation segmentation model to aggregate commonalities and introduce 2D prompts to learn peculiarities along the domain and task dimensions, respectively. 
\item A prompt discrimination loss is proposed to effectively enhance the learning of peculiarities and commonalities for 2D prompts and the foundation model, respectively. 
\item Our VSCode model surpasses all existing state-of-the-art models across all tasks on 26 datasets and showcases its ability to generalize to unseen tasks, further emphasizing the superiority of our approach.
\end{compactitem}

\vspace{-3mm}
\section{Related Work}
\label{sec:relatedwork}

\vspace{-1mm}
\subsection{Deep Learning Based SOD and COD}
\vspace{-1mm}
\textbf{SOD.}
Previous RGB SOD works delved into attention-based \cite{Piao2019dmra,zhang2018pagr,fan2020bbsnet,chen2020dpanet,liu2018picanet}, multi-level fusion-based \cite{hou2018dss,wang2017stagewise,MINet-CVPR2020,GateNet,fang2022densely, zhang2019synthesizing}, recurrent-model-based \cite{liu2016dhsnet,wang2018rfcn,deng2018r3net,liu2019salient,chen2020PGAR}, and multi-task-based methods \cite{wang2018salient,zhang2019capsal,qin2019basnet, zhao2019EGNet,CVPR2020_LDF,zhang2020select,Wei2020CoNet}. 
%As for multimodal SOD tasks, effectively exploring and integrating complementary cues is crucial. 
In the case of RGB-D SOD, some models \cite{zhao2019contrast, li2020icnet, Piao2019dmra, Li2020CMWNet, liu2020S2MA, liu2021learning, chen2020dpanet} leveraged various attention mechanisms to incorporate depth cues into RGB features.
With regard to RGB-T SOD, recent studies also introduced attention-based methods \cite{tu2022rgbt,wang2021cgfnet} and multi-level fusion \cite{zhang2019rgb, tu2021multi} to excavate the relationship between RGB and thermal features.
Regarding the VSOD task, 
some works \cite{wang2017video,ji2020casnet,song2018pyramid,fan2019shifting,gu2020pyramid} mined spatial-temporal and appearance cues.
%from consecutive RGB frames.
%via convolution networks. 
More recently, there was a growing trend where various research \cite{ren2020tenet,li2018flow,ji2023full,liu2023learning} endeavored to incorporate optical flow for combining motion cues with appearance details.
%from RGB frames. 
Consistent with recent studies, we treat optical flow as a form of modality information and view VSOD as a multimodal SOD task.

\vspace{1mm}
\noindent
\textbf{COD.}
Currently, COD has RGB COD, RGB-D COD, and VCOD tasks. RGB COD methods can be broadly categorized as multi-task-based approaches \cite{lv2021simultaneously,zhai2021mutual}, multi-input-based approaches \cite{pang2022zoom,zheng2023mffn}, and refinement-based approaches \cite{fan2020camouflaged,jia2022segment}. The RGB-D COD task was initially introduced in \cite{wu2023source}, where depth inference models are adapted for object segmentation. For VCOD, 
%motion cues hold greater importance than visual cues. 
prior studies segmented the moving camouflaged objects via dense optical flow \cite{bideau2016s,bideau2018moa} or well-designed models \cite{lamdouar2020betrayed,cheng2022implicit}. For a more comprehensive literature review, please refer to \cite{fan2023advances}.

\vspace{-4mm}
\subsection{Prompt in Computer Vision}
\vspace{-1mm}
Prompt was initially introduced in the field of NLP \cite{brown2020language} and has been successfully integrated into computer vision tasks \cite{hu2023compositional}.
%, yielding impressive outcomes. 
%In some works \cite{bahng2022exploring,huang2023diversity}, prompts were defined as pixels and added to each image. 
VPT \cite{jia2022visual} introduced a small amount of trainable parameters as prompts in the input space. ViPT \cite{zhu2023visual} put forth the idea of modality-complementary prompts for task-oriented multi-modal tracking.
Prior research has primarily focused on specific tasks, such as classification or tracking. In this paper, we propose to use 2D prompts for assembling different multimodal tasks and enable zero-shot generalization on unseen tasks, which has not been explored before.

\begin{figure*}[!t]
    \centering
    \captionsetup{skip=5pt}
    \includegraphics[width=1\linewidth]{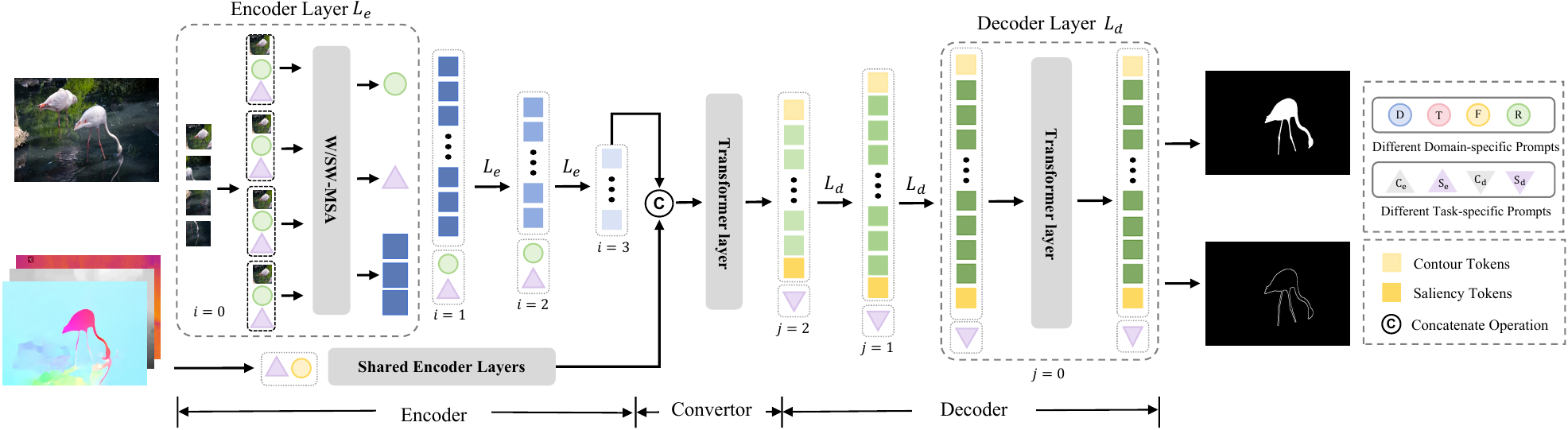}
    \caption{\textbf{Overall framework of our proposed VSCode model with 2D prompt learning.} Based on the VST \cite{Liu_2021_ICCV} foundation model, we insert the respective domain-specific prompts and task-specific prompts in the attention windows in the Swin transformer \cite{liu2021Swin} encoder layers to learn domain and task-specific encoder features. The convertor is used for multimodal feature fusion. Within the transformer decoder layers, task-specific prompts are appended to image feature tokens to perform task-specific decoding. We also provide detailed structures of an encoder layer ($i=0$) and a decoder layer  ($j=0$).}
    \label{method}
    \vspace{-4mm}
\end{figure*}

\vspace{-1.5mm}
\subsection{Generalist Segmentation Architecture}
\vspace{-1.5mm}
Recently, several generalist frameworks have emerged for a range of segmentation tasks using a variety of prompts \cite{awais2023foundational}.
On one hand, 
X-Decoder \cite{zou2023generalized} utilized generic non-semantic queries and semantic queries to decode different pixel-level and token-level outputs. UNINEXT \cite{yan2023universal} introduced three types of prompts, namely category names, language expressions, and reference annotations.
On the other hand, Painter \cite{wang2023images} and SegGPT \cite{wang2023seggpt} leveraged image-mask pairs from the same task as prompts. 
%to assist the model in executing individual segmentation tasks.
Unlike the approaches mentioned above, which mainly concentrate on task differences, our VSCode dissects unique characteristics based on both domain and task dimensions, leading to a more versatile design. 

In the field of SOD and COD, EVP \cite{liu2023explicit} introduced adaptors into the encoder and trained each task individually for various foreground segmentation tasks.  
Different from them, we consider not only multiple tasks but also multiple modalities and we train all tasks simultaneously.
%. Moreover, our VSCode trains all tasks simultaneously, leading to more sufficient data usage and efficient computational costs, which further distinguish it from EVP.

\vspace{-1mm}
\section{Methodology}
\vspace{-1mm}
In this work, we propose VSCode with the aim of jointly training SOD and COD tasks in an efficient and effective way. We allow VST \cite{Liu_2021_ICCV} to incorporate commonalities (Section~\ref{sec:foudation}), and utilize 2D prompts, which comprise domain-specific (Section~\ref{sec:domain-specific}) and task-specific prompts (Section~\ref{sec:task-specific}), to encapsulate peculiarities. 
To accurately disentangle domain and task peculiarities in 2D prompts and encourage commonality learning in VST, we introduce a prompt discrimination loss (Section~\ref{sec:loss}). Figure~\ref{method} shows the overall architecture of our proposed VSCode.
\label{sec:method}

\vspace{-1mm}
\subsection{Foundation Model}
\vspace{-1mm}
\label{sec:foudation}
To achieve a more comprehensive integration of commonalities from SOD and COD tasks, we select VST \cite{Liu_2021_ICCV} as our fundamental model. VST was originally proposed for RGB and RGB-D SOD and comprises three primary components, \ie\ a transformer encoder, a transformer convertor, and a multi-task transformer decoder. It initially employs the transformer encoder to capture long-range dependencies within the image features $\bm{f}_i^E \in{\mathbb{R}^{l_i \times c_i}}$, where $i \in [0, 1, 2, 3]$ indicates the index of blocks in the encoder, $l_i$ and $c_i$ mean the length of the patch sequence and the channel number of $\bm{f}_i^E$. Subsequently, the transformer convertor integrates the complement between RGB and depth features via cross-attention for RGB-D SOD or uses self-attention for RGB SOD. In the decoder, which is composed of a sequence of self-attention layers, VST predicts saliency maps and boundary maps simultaneously via a saliency token, a boundary token, and decoder features $\bm{f}_j^D \in{\mathbb{R}^{l_j \times d}}$, where $j$ corresponds the index of blocks in the decoder. Here $j \in [2,1,0]$
for descending order and $d$ =384. Due to the simple and pure-transformer-based architecture, VST can be easily used for other multimodal tasks and COD tasks without the need for model redesign. As a result, it emerges as a superior choice for constructing a generalist model for general multimodal SOD and COD.

In pursuit of improved outcomes and a more suitable structure, we introduce modifications to VST. First, we select Swin transformer \cite{liu2021Swin} as our backbone due to its efficiency and high performance. 
Second, to maintain a unified structure for both RGB tasks and other multimodal tasks, we utilize the RGB convertor in VST, which comprises standard transformer layers. For multimodal tasks, we simply concatenate the supplementary modality's features with the RGB features along the channel dimension and employ a multilayer perceptron (MLP) to project them from 2$d$ channels to $d$ channels. For RGB tasks, no alterations are made. Third, we incorporate certain extensions from VST\textbf{++} \cite{liu2023vst++}, specifically including the token-supervised prediction loss.

\vspace{-1mm}
\subsection{Domain-specific Prompt}
\vspace{-1mm}
\label{sec:domain-specific} 
Within the encoder, lower layers are dedicated to extracting low-level features, encompassing edges, colors, and textures, which exhibit distinct characteristics in various domains \cite{zeiler2014visualizing}. For instance, depth maps are typically rendered in grayscale, while thermal maps present a broader color spectrum. Higher layers, on the other hand, capture semantic information from modality features, which is crucial for all tasks. Consequently, we introduce domain-specific prompts $\bm{p}_i^{d}$ at each block $i$ in the encoder and design four kinds of domain-specific prompts for RGB, depth, thermal, and optical flow, respectively, to highlight the disparities among domains, as shown in Figure~\ref{method}. 

Given the image features $\bm{f}_i^{E}$ from a specific block in the Swin transformer encoder,
we use window-attention \cite{liu2021Swin} and partition the feature $\bm{f}_i^{E}$ into window features $\bm{f}_{i\_w}^{E} \in{\mathbb{R}^{l_i/M^2 \times M^2 \times c_i}}$, where $M$ represents the window size and $l_i/M^2$ is the number of windows. Then, we replicate the prompts $\bm{p}_i^{d} \in{\mathbb{R}^{N_i\times c_i}}$ for each window and obtain $\bm{p}_i^{d'} \in{\mathbb{R}^{l_i/M^2 \times N_i\times c_i}}$, where $N_i$ represents the number of learnable prompt tokens. Next, we append them to the patch feature tokens in each window and perform self-attention within each window, which can be defined as
\begin{equation} \label{CFP-window}
\begin{bmatrix}
\bm{p}_{i+1}^{d'} \\
\bm{f}_{i\_w}^{E}
\end{bmatrix}
\leftarrow \text{MLP}(\text{SW/W-MSA}(
\begin{bmatrix}
\bm{p}_{i}^{d'} \\
\bm{f}_{i\_w}^{E}
\end{bmatrix}
)),
\end{equation}
where W-MSA and SW-MSA are multi-head self-attention modules with regular and shifted windowing configurations, respectively.
Here we omit the residual connection \cite{he2016resnet}, and layer normalization \cite{ba2016layer}.
Next, we segment $\bm{p}_{i+1}^{d'}$ from each window and calculate the average of them to obtain $\bm{p}_{i+1}^{d}$, and then reassemble the output window feature $\bm{f}_{i\_w}^{E}$ to $\bm{f}_{i+1}^{E}$
for the next block.

\vspace{-1mm}
\subsection{Task-specific Prompt}
\begin{figure}[!t]
    \centering
    \captionsetup{skip=17pt}
    \begin{overpic}[width=1\linewidth]{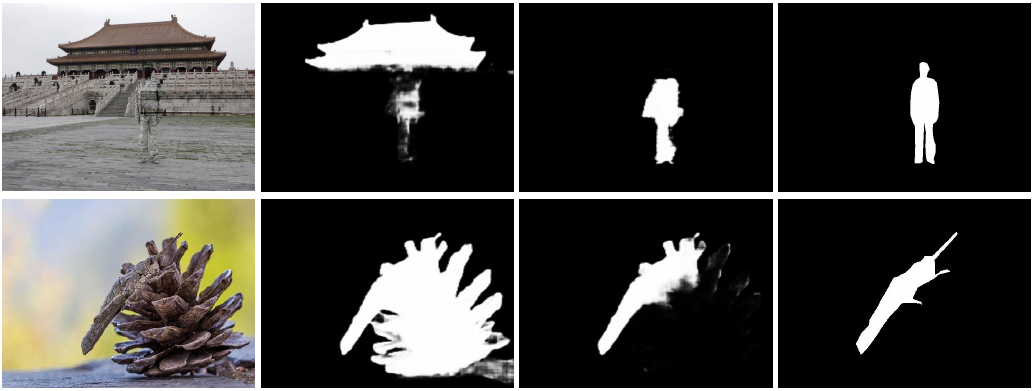}
    \put(7.7,-4.5){\footnotesize Image}
    \put(27.6,-6.5){\scriptsize \shortstack[c] {Predictions with \\SOD Prompt}}
    \put(53.3,-6.5){\scriptsize \shortstack[c] {Predictions with \\COD Prompt}}
    \put(82.3,-4.5){\footnotesize COD GT}
    \end{overpic}
    \caption{\textbf{Illustration of the influence of using different task prompts.}}
    \label{example}
    \vspace{-0.5cm}
\end{figure}
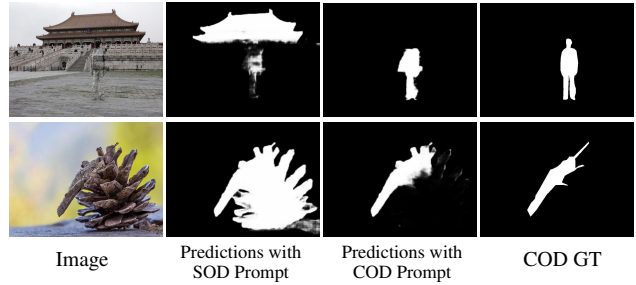

\vspace{-1mm}
\label{sec:task-specific}
Prior research \cite{li2021uncertainty} has traditionally regarded SOD and COD as opposing tasks, emphasizing the disparities between the features extracted by the SOD encoder and the COD encoder as much as possible. However, we believe that SOD and COD share significant commonalities in their features, such as low-level cues, high-level objectness, and spatial structuredness. As a result, we introduce task-specific prompts to learn the peculiarities while retaining the primary stream parameters shared to capture commonalities. We add the task-specific prompts in both VST encoder and decoder, and the overall impact of adding these prompts is illustrated in Figure~\ref{example}.

%\vspace{2mm}
\textbf{Encoder.} 
Although the encoder primarily focuses on domain-specific features with domain prompts, semantic features still play a pivotal role in distinguishing SOD and COD tasks.  
Semantic features from the encoder typically emphasize the most relevant region for a particular task and allocate more attention accordingly.
In the case of the SOD task, the foreground region receives greater attention, whereas for the COD task, the background usually gains large importance since objects are typically concealed within it.
Hence, it is essential to incorporate task-specific prompts to encourage learning task-related features in the encoder. 
Otherwise, we risk initially activating the wrong objects before the decoding process. 
Following the pattern of domain-specific prompts, we introduce task-specific prompts $\bm{p}_i^{te} \in \mathbb{R}^{N_i\times c_i}$ in each encoder block and use them in the same way as how domain-specific prompts are used.

%\vspace{2mm}
\textbf{Decoder.}
Camouflaged objects typically exhibit more intricate and detailed boundaries compared to salient objects. This complexity arises because concealed objects often share color or textual similarities with their surroundings, resulting in imperceptible boundaries. Therefore, solely introducing task-specific prompts in the encoder may not be adequate, as camouflaged objects require a more refined process within the decoder. We incorporate task-specific prompts in the decoder to allocate distinct attention for reconstructing both the boundary and object regions based on the features extracted by the encoder.
In contrast, previous research \cite{li2021uncertainty} has not adequately explored the differences between these two tasks in the decoder, as they typically use a single decoder to handle both.

Regarding task-specific prompts in the decoder, 
we simply append learnable prompts $\bm{p}_{j+1}^{td} \in \mathbb{R}^{N\times d}$ to the decoder feature tokens $\bm{f}_{j+1}^{D}$ from a specific block $j+1$ in the decoder. Then, we apply the self-attention as follows:
\begin{equation} \label{taskp}
\begin{bmatrix}
\bm{p}_{j}^{td} \\
\bm{f}_{j}^{D}\\
\end{bmatrix}
\leftarrow \text{MLP}(\text{MSA}(
\begin{bmatrix}
\bm{p}_{j+1}^{td} \\
\bm{f}_{j+1}^{D}\\
\end{bmatrix}
)),
\end{equation}
where MSA denotes the multi-head self-attention. Here we omit the saliency and boundary tokens in the VST decoder for conciseness. 
Please note that our task-specific prompts differ from saliency and boundary tokens since we do not introduce any supervision for them.

\vspace{-2mm}
\subsection{Prompts Layout and Discussion}
\vspace{-1mm}
\label{sec:discus}
To incorporate the aforementioned prompts within the encoder-decoder architecture, inspired by VPT \cite{jia2022visual}, we offer two prompt inserting versions. In the deep version, new prompts are introduced at the start of each transformer block, whereas the shallow version involves proposing prompts at first and updating them across all blocks.
To unveil the specific relationship among different domains and tasks at varying network depths, we employ the deep version for both domain and task-specific prompts within the encoder. Based on VST's design, which introduces a saliency and a boundary token at the beginning of the decoder, we use the shallow version for task-specific prompts in the decoder.

\begin{figure}[!t]
    \centering
    \captionsetup{skip=3pt}
    \includegraphics[width=1\linewidth]{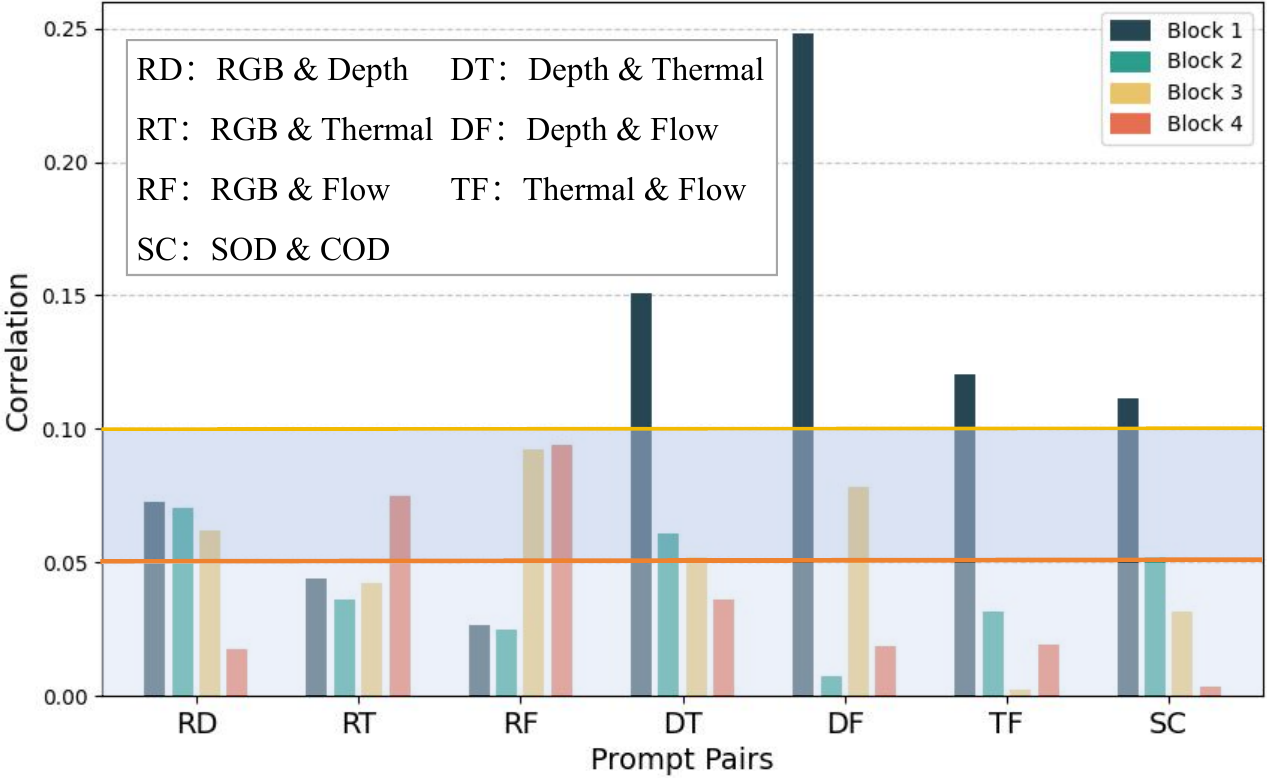}
    \caption{\textbf{Correlation of prompt pairs at each encoder block.}}
    \label{relation}
    \vspace{-0.5cm}
\end{figure}

We calculate the correlations of different domain and task prompt pairs at different blocks in Figure~\ref{relation}. It is evident that depth, thermal, and optical flow exhibit relatively strong correlations in low-level features, 
as all of them usually show obvious low-level contrast between target objects and backgrounds in terms of color or luminance.
However, at higher levels, most domains exhibit lower correlations, highlighting the distinctions among them.
Additionally, as for task-specific prompts, it is clear that SOD prompts and COD prompts exhibit more shared knowledge in the lower layers. As we progress to higher layers, the correlation decreases, indicating that high-level features gradually learn unrelated information.  
This observation urges us to implement the deep version of domain-specific prompts and task-specific prompts in the encoder in our final design, as different blocks acquire distinct knowledge. 
Moreover, the gradually decreased correlation values along with the increase of the network depth encourage us to use a progressively larger number of prompt tokens, as lower correlation means larger peculiarities and hence requires more parameters to learn.

\vspace{-1mm}
\subsection{Loss Function}
\vspace{-1mm}
\label{sec:loss} 
The design principle of our model is to use 2D prompts for encompassing peculiarities while integrating commonalities into the foundation model. 
However, this is not straightforward for freely learned prompts.
As shown in Figure~\ref{relation}, they still suggest certain correlations. This indicates that the learned prompts are entangled, risking the model's capacity to differentiate among various domains and tasks and resulting in suboptimal optimization.
Hence, we propose a prompt discrimination loss to minimize the correlation among the prompts of the same type, guaranteeing that each prompt acquires unique domain or task knowledge. 
Specifically, we aggregate prompts of the same domain/task into a single embedding and then perform discrimination.
First, we average the input prompt tokens of each same prompt type at each block
and use linear projections to align the channel numbers to $d$. 
Subsequently, for each type of prompt, we concatenate the averaged prompts of different blocks, and use MLP to obtain the overall domain-specific prompt $p_{l}^{d_{all}}$ and task-specific encoder prompt $p_{k}^{{te}_{all}}$:
\begin{equation} \label{aggregateprompt}
\begin{split}
p_{l}^{d_{all}} = \text{MLP}[\text{LA}(\bm{p}_0^{d});\text{LA}(\bm{p}_1^{d});\text{LA}({\bm{p}_2^{d}});\text{LA}(\bm{p}_3^{d})],\\
p_{k}^{{te}_{all}} = \text{MLP}[\text{LA}(\bm{p}_0^{te});\text{LA}(\bm{p}_1^{te});\text{LA}({\bm{p}_2^{te}});\text{LA}(\bm{p}_3^{te})],
\end{split}
\end{equation}
where L and A represent the linear and average operation, respectively, with $l \in \{depth, thermal, flow, rgb\}$, and $k \in \{SOD, COD\}$. Since the task-specific prompts in the decoder are shallow, we simply average them.

Afterward, we calculate the cosine similarity between prompt pairs,
resulting in eight types of cosine similarity results $\mathcal{CS}_{m}$. Here $m$ means the combination of domains/tasks, namely $\{RD, RT, RF, DF, DT, TF\}$ for domain-aggregated prompts and $\{SC_{EN}, SC_{DE}\}$ for task-aggregated prompts in the encoder and decoder, respectively. 
Finally, we minimize the correlation within these prompt pairs
and define our prompt discrimination loss as
\begin{equation} \label{cosinloss}
\mathcal{L}_{dis} = \sum_{m}\ln(1+|\mathcal{CS}_{m}|),
\end{equation}
which is further combined with the segmentation losses and boundary losses \cite{Liu_2021_ICCV} to train our model.

\vspace{-2mm}
\section{Experiment}

 %%%%%%%%% Ablation study all
\begin{table*}[t]
  \centering
  %\scriptsize
  \footnotesize
  \renewcommand{\arraystretch}{1.0}
\setlength\tabcolsep{0.45mm}
\vspace{-4mm}
%\begin{adjustbox}{width=\linewidth}
 \begin{tabular}{c|c|ccc|ccc|ccc|ccc|ccc|ccc}
  \toprule[1pt]
  \multicolumn{1}{c|}{\multirow{3}{*}{Settings}} &\multicolumn{1}{c|}{\multirow{2}{*}{Params}}  & \multicolumn{3}{c|}{\textbf{RGB SOD}}  & \multicolumn{3}{c|}{\textbf{RGB-D SOD}}  & \multicolumn{3}{c|}{\textbf{RGB-T SOD}} & \multicolumn{3}{c|}{\textbf{VSOD}} & \multicolumn{3}{c|}{\textbf{RGB COD}} & \multicolumn{3}{c}{\textbf{VCOD}}\\
  \multicolumn{1}{c|}{} & \multicolumn{1}{c|}{}& \multicolumn{3}{c|}{DUTS\cite{wang2017duts}} & \multicolumn{3}{c|}{NJUD\cite{ju2014njud}} & \multicolumn{3}{c|}{VT5000\cite{tu2022rgbt}}  & \multicolumn{3}{c|}{SegV2\cite{li2013video}} & \multicolumn{3}{c|}{CAMO\cite{le2019anabranch}} & \multicolumn{3}{c}{CAD\cite{bideau2016s}}\\
   {} & {(M)} & $S_m\uparrow$ & $F_m\uparrow$ & $E_m\uparrow$  & $S_m\uparrow$ & $F_m\uparrow$ & $E_m\uparrow$ & $S_m\uparrow$ & $F_m\uparrow$ & $E_m\uparrow$ & $S_m\uparrow$ & $F_m\uparrow$ & $E_m\uparrow$ & $S_m\uparrow$ & $F_m\uparrow$ & $E_m\uparrow$ & $S_m\uparrow$ & $F_m\uparrow$ & $E_m\uparrow$ \\ \hline
   {ST} &{323.46\footnotemark[1]} &.900 &.885 &.940  &.927 &.928 &.958 &.900 &.863 &.938  &.896 &.870 &.952 &.793 &.751 &.871 &.686 &.522 &.787\\\hline
   {GT} &{54.06} &.898 &.884 &.941 &.924 &.922 &.954 &.903 &.886 &.942 &.930 &.911 &.972 &- &- &-  &- &- &-\\
   {GT+$p^d$} &{54.06} &.902 &.890 &.945 &.931 &.932 &.962 &.909 &.877 &.947 &.931 &.917 &.975 &- &- &-  &- &- &- \\
   {GT+$p^d$+$p^t$} &{54.09} &.904 &.892 &.945 &.931 &.931 &.961 &.906 &\textbf{.892} &.946 &.925 &.910 &.970 &.804 &.776 &.876 &\textbf{.759} &\textbf{.639} &.808\\
   \rowcolor{c1!50}{\textbf{GT+$\bm{p^d}$+$\bm{p^t}$+$\bm{\mathcal{L}_{dis}}$}} &{54.09} &\textbf{.909} &\textbf{.899} &\textbf{.948} &\textbf{.935} &\textbf{.938} &\textbf{.965} &\textbf{.912} &.882 &\textbf{.950} &\textbf{.943} &\textbf{.930} &\textbf{.984}  &\textbf{.811} &\textbf{.782} &\textbf{.884} &.736 &.614 &.797\\\hline
   {w/o $p^{te}$} &{54.07} &.908 &.896 &.947 &.932 &.932 &.960 &.909 &.878 &.947 &.933 &.907 &.966 &.800 &.770 &.872  &.743 &.611 &.798 \\
   {w/o $p^{td}$} &{54.08} &.902 &.889 &.943 &.929 &.929 &.959 &.904 &.872 &.941 &.940 &.919 &.975 &.799 &.770 &.875  &.740 &.599 &\textbf{.814} \\ 
   \bottomrule[1pt]
   \end{tabular}
   \vspace{-3.5mm}
    \caption{\textbf{Ablation studies of our VSCode on the Swin-T \cite{liu2021Swin} backbone with $224 \times 224$ image size.} We conduct evaluations on one representative dataset for each task. 
    ``ST'' indicates special training, ``GT'' means general training, $p^d$ represents domain-specific prompts, and $p^t$ is task-specific prompts, which consists of $p^{te}$ in the encoder and $p^{td}$ in the decoder. $\mathcal{L}_{dis}$ is our prompt discrimination loss. 
 The best results under each setting are labeled in \textbf{bold}.}
   \vspace{1mm}
  \label{ALL_ablationstudy}
 \end{table*}

 %%%%%%%%% Ablation study each part
\begin{table*}[t]
  \centering
  %\scriptsize
  \footnotesize
  %\small
  \renewcommand{\arraystretch}{1}
  \renewcommand{\tabcolsep}{0.8mm}
\vspace{-2mm}
%\begin{adjustbox}{width=\linewidth}
 \begin{tabular}{c|c|ccc|ccc|ccc|ccc|ccc|ccc}
  \toprule[1pt]
  \multicolumn{1}{c|}{\multirow{3}{*}{Settings}} &\multicolumn{1}{c|}{\multirow{2}{*}{Params}}  & \multicolumn{3}{c|}{\textbf{RGB SOD}}  & \multicolumn{3}{c|}{\textbf{RGB-D SOD}}  & \multicolumn{3}{c|}{\textbf{RGB-T SOD}} & \multicolumn{3}{c|}{\textbf{VSOD}} & \multicolumn{3}{c|}{\textbf{RGB COD}} & \multicolumn{3}{c}{\textbf{VCOD}}\\
  %\cline{-25}
  \multicolumn{1}{c|}{} & \multicolumn{1}{c|}{}& \multicolumn{3}{c|}{DUTS\cite{wang2017duts}} & \multicolumn{3}{c|}{NJUD\cite{ju2014njud}} & \multicolumn{3}{c|}{VT5000\cite{tu2022rgbt}}  & \multicolumn{3}{c|}{SegV2\cite{li2013video}} & \multicolumn{3}{c|}{CAMO\cite{le2019anabranch}} & \multicolumn{3}{c}{CAD\cite{bideau2016s}}\\
  {} & {(M)} & $S_m\uparrow$ & $F_m\uparrow$ & $E_m\uparrow$  & $S_m\uparrow$ & $F_m\uparrow$ & $E_m\uparrow$ & $S_m\uparrow$ & $F_m\uparrow$ & $E_m\uparrow$ & $S_m\uparrow$ & $F_m\uparrow$ & $E_m\uparrow$ & $S_m\uparrow$ & $F_m\uparrow$ & $E_m\uparrow$ & $S_m\uparrow$ & $F_m\uparrow$ & $E_m\uparrow$ \\ \hline
   \multicolumn{20}{l}{\textbf{shallow or deep version for domain-specific prompts}}\\
   shallow &54.83 &.900 &.887 &.943 &.926 &.926 &.955 &.905 &.873 &.944 &\textbf{.931} &\textbf{.917} &.972 &- &- &-  &- &- &- \\ 
   \rowcolor{c1!50}\textbf{deep} &{54.06} &\textbf{.902} &\textbf{.890} &\textbf{.945} &\textbf{.931} &\textbf{.932} &\textbf{.962}  &\textbf{.909} &\textbf{.877} &\textbf{.947}  &\textbf{.931} &\textbf{.917} &\textbf{.975} &- &- &- &- &- &-\\\hline
   \multicolumn{20}{l}{\textbf{shallow or deep version for task-specific prompts in the encoder}}\\
   shallow  &54.45 &.902 &.888 &.943 &.928 &.928 &.958 &.902 &.870 &.940 &\textbf{.927} &\textbf{.905} &\textbf{.964} &.793 &.763 &.866 &.747 &.616 &.798\\
  \rowcolor{c1!50} \textbf{deep} &{54.08} &\textbf{.903} &\textbf{.890} &\textbf{.944} &\textbf{.934} &\textbf{.934} &\textbf{.962} &\textbf{.905} &\textbf{.874} &\textbf{.943} &.924 &.903 &.960 &\textbf{.804} &\textbf{.772} &\textbf{.881} &\textbf{.759} &\textbf{.651} &\textbf{.831} \\\hline
   \multicolumn{20}{l}{\textbf{shallow or deep version for task-specific prompts in the decoder}}\\
   deep &54.10 &.903 &.891 &\textbf{.945} &\textbf{.930} &\textbf{.932} &.960 &.905 &.888 &.943 &.922 &.903 &.966 &.802 &.774 &\textbf{.881} &.738 &.605 &.801 \\
   \rowcolor{c1!50}\textbf{shallow} &{54.09} &\textbf{.904} &\textbf{.892} &\textbf{.945} &\textbf{.930} &.931 &\textbf{.961} &\textbf{.906} &\textbf{.892} &\textbf{.946} &\textbf{.925} &\textbf{.910} &\textbf{.970} &\textbf{.804} &\textbf{.776} &.876  &\textbf{.759} &\textbf{.639} &\textbf{.808} \\\hline
   \multicolumn{20}{l}{\textbf{number of domain-specific prompts at four blocks}}\\
   \rowcolor{c1!50} \textbf{1,1,1,1} &{54.06} &.902 &\textbf{.890} &\textbf{.945} &\textbf{.931} &.932 &\textbf{.962}  &\textbf{.909} &\textbf{.877} &\textbf{.947}  &\textbf{.931} &\textbf{.917} &\textbf{.975} &- &- &- &- &- &-\\
   5,5,5,5 &54.08 &\textbf{.903} &.890 &\textbf{.945} &.931 &\textbf{.935} &.961 &.902 &.869 &.940 &.918 &.887 &.954 &- &- &- &- &- &-\\\hline
   \multicolumn{20}{l}{\textbf{number of task-specific prompts in the encoder at four blocks}}\\
   %1:1:1:1 &54.50 &\textbf{0.903} &0.890 &0.944 &0.933 &0.934 &0.962 &\textbf{0.908} &\textbf{0.876} &\textbf{0.945} &\textbf{0.939} &\textbf{0.924} &\textbf{0.977} &0.676 &0.612 &0.740 &0.737 &0.599 &0.763\\ 
   5,5,5,5 &54.07 &\textbf{.903} &\textbf{.893} &\textbf{.947} &.928 &.930 &.959  &.903 &.870 &.940 &\textbf{.931} &\textbf{.918} &\textbf{.975} &.795 &.766 &.866 &.739 &.600 &.799 \\
   \rowcolor{c1!50} \textbf{1,1,5,10} &{54.08} &\textbf{.903} &.890 &.944 &\textbf{.934} &\textbf{.934} &\textbf{.962} &\textbf{.905} &\textbf{.874} &\textbf{.943} &.924 &.903 &.960 &\textbf{.804} &\textbf{.772} &\textbf{.881} &\textbf{.759} &\textbf{.651} &\textbf{.831} \\\hline
   \multicolumn{20}{l}{\textbf{number of task-specific prompts in the decoder}}\\
   5 &54.08 &\textbf{.904} &.890 &\textbf{.946} &.929 &.931 &.957 &.904 &.890 &.943 &.931 &.911 &.969 &\textbf{.807} &\textbf{.782} &\textbf{.881}  &.746 &.626 &.805 \\
   \rowcolor{c1!50} \textbf{10} &{54.09} &\textbf{.904} &\textbf{.892} &.945 &\textbf{.930} &.931 &\textbf{.961} &\textbf{.906} &\textbf{.892} &\textbf{.946} &.925 &.910 &.970 &.804 &.776 &.876 &\textbf{.759} &\textbf{.639} &\textbf{.808} \\
   15 &54.09 &.903 &.889 &.944 &.929 &\textbf{.933} &.956 &.904 &.890 &.942 &\textbf{.932} &\textbf{.913} &\textbf{.974} &.798 &.771 &.875  &.743 &.621 &.791\\
  \bottomrule[1pt]
   \end{tabular}
   \vspace{-3.5mm}
   \caption{\textbf{Ablation studies of different designs of prompt layout.} 
 }
  \label{eachpartablationstudy}
  \vspace{-4.5mm}
 \end{table*}

 \vspace{-1mm}
\subsection{Datasets and Evaluation Metrics}
For \colorbox{gray!20}{RGB SOD}, we evaluate our proposed model using six commonly used benchmark datasets, \ie\ \textbf{DUTS} \cite{wang2017duts}, \textbf{ECSSD} \cite{yan2013ECSSD}, \textbf{HKU-IS} \cite{li2015HKUIS}, \textbf{PASCAL-S} \cite{li2014PASCALS}, \textbf{DUT-O} \cite{yang2013DUTO}, and \textbf{SOD} \cite{movahedi2010SOD}. 
For \colorbox{gray!20}{RGB-D SOD}, we use six large benchmark datasets, including \textbf{STERE} \cite{niu2012stere}, \textbf{NJUD} \cite{ju2014njud}, \textbf{NLPR} \cite{peng2014nlpr}, \textbf{DUTLF-Depth} \cite{Piao2019dmra}, \textbf{SIP} \cite{fan2020SIP}, and \textbf{ReDWeb-S} \cite{liu2021learning}.
In terms of \colorbox{gray!20}{RGB-T SOD}, we consider three public datasets: \textbf{VT821} \cite{wang2018rgb}, \textbf{VT1000} \cite{tu2019rgb}, and \textbf{VT5000} \cite{tu2022rgbt}. 
For \colorbox{gray!20}{VSOD}, we employ six widely used benchmark datasets: \textbf{DAVIS} \cite{perazzi2016benchmark}, \textbf{FBMS} \cite{ochs2013segmentation}, \textbf{ViSal} \cite{wang2015consistent}, \textbf{SegV2} \cite{li2013video}, \textbf{DAVSOD-Easy}, and \textbf{DAVSOD-Normal} \cite{fan2019shifting}.
Regarding \colorbox{gray!20}{RGB COD}, three extensive benchmark datasets are considered, including \textbf{COD10K} \cite{fan2020camouflaged}, \textbf{CAMO} \cite{le2019anabranch}, and \textbf{NC4K} \cite{lv2021simultaneously}.
For \colorbox{gray!20}{VCOD}, we utilize two widely accepted benchmark datasets: \textbf{CAD} \cite{bideau2016s} and \textbf{MoCA-Mask} \cite{cheng2022implicit}.
\footnotetext[1]{The parameters for our specialized training methods amount to 53.61M for the RGB task and 54.06M for the multimodal task, resulting in a total of 323.46M parameters for all six tasks.}
To ensure a consistent evaluation across all SOD and COD tasks, we employ three commonly used evaluation metrics to assess model performance: structure-measure $S_m$ \cite{fan2017structure}, maximum enhanced-alignment measure $E_m$ \cite{Fan2018Enhanced}, and maximum F-measure $F_m$.

\vspace{-1mm}
\subsection{Implementation Details}
\vspace{-1mm}
Building on prior research \cite{Liu_2021_ICCV,tu2022rgbt,liu2023learning,he2023camouflaged,cheng2022implicit}, we employ the following datasets to train our model concurrently: the training set of \textbf{DUTS} for \colorbox{gray!20}{RGB SOD}, the training sets of \textbf{NJUD}, \textbf{NLPR}, and \textbf{DUTLF-Depth} for \colorbox{gray!20}{RGB-D SOD}, the training set of \textbf{VT5000} for \colorbox{gray!20}{RGB-T SOD}, the training sets of \textbf{DAVIS} and \textbf{DAVSOD} for \colorbox{gray!20}{VSOD}, the training sets of \textbf{COD10K} and \textbf{CAMO} for \colorbox{gray!20}{RGB COD}, and the training set of \textbf{MoCA-Mask} for \colorbox{gray!20}{VCOD}.
To ensure a fair comparison with previous works \cite{zhuge2021salient,lee2022spsn,tu2021multi,liu2023explicit,he2023camouflaged}, 
we resize each image to $384 \times 384$ pixels and then randomly crop them to $352 \times 352$ image regions for training.
Our training process employs the Adam optimizer \cite{kingma2014adam}
with an initial learning rate of 0.0001, which is reduced by a factor of 10 at half and three-quarters of the total training steps. We conduct a total of 150,000 training steps using a 3090 GPU. 
We mix the above six tasks in each training iteration with two samples for each task, leading to a total batch size of 12.

 %%%%%%%%% RGB SOTA
\begin{table*}[t]
  \centering
  %\scriptsize
  \footnotesize
  %\small
  \renewcommand{\arraystretch}{1}
  \renewcommand{\tabcolsep}{0.65mm}
\vspace{-2mm}
%\begin{adjustbox}{width=\linewidth}
 \begin{tabular}{l|c|ccc|ccc|ccc|ccc|ccc|ccc}
  \toprule[1pt]
  \multicolumn{1}{c|}{\multirow{2}{*}{Method}} & \multicolumn{1}{c|}{\multirow{1}{*}{Params}} & \multicolumn{3}{c|}{DUTS\cite{wang2017duts}} & \multicolumn{3}{c|}{ECSSD\cite{yan2013ECSSD}} & \multicolumn{3}{c|}{HKU-IS\cite{li2015HKUIS}} & \multicolumn{3}{c|}{PASCAL-S\cite{li2014PASCALS}} & \multicolumn{3}{c|}{DUT-O\cite{yang2013DUTO}} & \multicolumn{3}{c}{SOD\cite{movahedi2010SOD}}\\
   {} &{(M)}
    & $S_m\uparrow$ & $F_m\uparrow$ & $E_m\uparrow$ & $S_m\uparrow$ & $F_m\uparrow$ & $E_m\uparrow$ & $S_m\uparrow$ & $F_m\uparrow$ & $E_m\uparrow$ & $S_m\uparrow$ & $F_m\uparrow$ & $E_m\uparrow$ & $S_m\uparrow$ & $F_m\uparrow$ & $E_m\uparrow$ & $S_m\uparrow$ & $F_m\uparrow$ & $E_m\uparrow$\\
   \hline
   %PiCANet\cite{liu2018picanet} &47.22 &0.863 &0.840 &0.915 &0.916 &0.929 &0.953 &0.905 &0.913 &0.951 &0.846 &0.824 &0.882 &0.826 &0.767 &0.865 &0.813 &0.824&0.871\\
   %AFNet\cite{Feng_AFNet} &35.95 &0.867 &0.838 &0.910 &0.914 &0.924&0.947 &0.905 &0.910 &0.949 &0.849 &0.824 &0.877 &0.826 &0.759 &0.861 &0.811 &0.819 &0.867 \\
   %TSPOANet\cite{Liu_TSPOANet} &- &0.860 &0.828 &0.907&0.907 &0.919 &0.942 &0.902 &0.909 &0.950 &0.841 &0.817 &0.871 &0.818 &0.750 &0.858 &0.802 &0.809 &0.852 \\
   %EGNet-R\cite{zhao2019EGNet} &111.64 &0.887 &0.866 &0.926 &0.925 &0.936 &0.955 &0.918 &0.923 &0.956 &0.852 &0.825 &0.874 &0.841 &0.778 &0.878 &0.824 &0.831 &0.875 \\
   %ITSD-R\cite{Zhou2020ITSD} &26.47 &0.885 &0.867 &0.929 &0.925 &0.939 &0.959 &0.917 &0.926 &0.960 &0.861 &0.839 &0.889 &0.840 &0.792 &0.880 &0.835 &0.849 &0.889\\
   %MINet-R\cite{MINet-CVPR2020} &162.38 &0.884 &0.864 &0.926 &0.925 &0.938 &0.957 &0.919 &0.926 &0.960 &0.856 &0.831 &0.883 &0.883 &0.769 &0.869  &0.830 &0.835&0.878\\
   %LDF-R\cite{CVPR2020_LDF} &25.15 &0.892 &0.877 &0.930 &0.925 &0.938 &0.954 &0.920 &0.929 &0.958 &0.861 &0.839 &0.888 &0.839 &0.782 &0.870 &0.831 &0.841 &0.878 \\
   %CSF-R2\cite{{gao2020sod100k}} &36.53 &0.890 &0.869 &0.929 &0.931 &0.942 &0.960 &- &- &- &0.863 &0.839 &0.885 &0.838 &0.775 &0.869 &0.826 &0.832 &0.883 \\
   %GateNet-R\cite{GateNet} &128.63 &0.891 &0.874 &0.932 &0.924 &0.935 &0.955 &0.921 &0.926 &0.959 &0.863 &0.836 &0.886 &0.840 &0.782 &0.878 &0.827 &0.835 &0.877 \\
   VST\cite{Liu_2021_ICCV}  &44.48 &.896 &.877 &.939 &.932 &.944 &.964  &.928 &.937 &{.968}  &{.873} &.850 &.900  &.850 &.800 &.888 &.854 &.866 &{.902} \\
   ICON-R\cite{zhuge2021salient} &33.09 &.890 &.876 &.931  &.928 &.943 &.960  &.920 &.931 &.960 &.862 &.844 &.888 &.845 &.799 &.884 &.848 &.861 &.899\\
   VST-T\texttt{++} \cite{liu2023vst++} &53.60 &.901 &.887 &.943 &{.937} &{.949} &{.968} &{.930} &{.939} &{.968} &{.878} &\textbf{.855} &\textbf{.901} &{.853} &{.804} &{.892}  &{.853} &{.866} &.899 \\
   MENet\cite{wang2023pixels} &27.83 &{.905} &{.895} &{.943}  &.927 &.938 &.956 &.927 &{.939} &.965 &.871 &.848 &.892  &.850 &.792 &.879  &.841 &.847 &.884 \\
   \rowcolor{c1!50}\textbf{VSCode-T} &{54.09} &\textbf{.917} &\textbf{.910} &\textbf{.954} &\textbf{.945} &\textbf{.957} &\textbf{.971} &\textbf{.935} &\textbf{.946} &\textbf{.970} &\textbf{.878} &{.852} &{.900} &\textbf{.869} &\textbf{.830} &\textbf{.910} &\textbf{.863} &\textbf{.879} &\textbf{.908} \\\hline
   EVP\cite{liu2023explicit} &64.52\footnotemark[2] &.917 &.910 &.956 &.936 &.949 &.965 &.935 &.945 &.971 &.880 &.859 &.902 &.864 &.822 &.902 &.854 &.873 &.901 \\
   \rowcolor{c1!50}\bes{VSCode-S} &{74.72\footnotemark[2]} &\bes{.926} &\bes{.922} &\bes{.960} &\bes{.949} &\bes{.959} &\bes{.974} &\bes{.940} &\bes{.951} &\bes{.974} &\bes{.887} &\bes{.864} &\bes{.904} &\bes{.877} &\bes{.840} &\bes{.912} &\bes{.870} &\bes{.882} &\bes{.910} \\
   \bottomrule[1pt]
   \end{tabular}
   \vspace{-3.5mm}
   \caption{\textbf{Quantitative comparison of our VSCode with other 5 SOTA RGB SOD methods on six benchmark datasets.}
``-R", ``-T" and ``-S"  mean the ResNet50 \cite{he2016resnet}, Swin-T, and Swin-S\cite{liu2021Swin} backbones, respectively. `-' indicates the code is not available.
The best performance under all settings is \bes{bolded}, and the best results under each setting are labeled in \textbf{bold}. }
\vspace{1mm}
  \label{RGB_SOTA}
 \end{table*}

%%%%%%%%% RGBD SOTA
 \begin{table*}[t]
  \centering
  %\scriptsize
  \footnotesize
  %\small
  \renewcommand{\arraystretch}{1.0}
  \renewcommand{\tabcolsep}{0.65mm}
\vspace{-2mm}
  %\resizebox{\linewidth}{!}{
  \begin{tabular}{l|c|ccc|ccc|ccc|ccc|ccc|ccc}
  \toprule[1pt]
  \multicolumn{1}{c|}{\multirow{2}{*}{Method}} & \multicolumn{1}{c|}{\multirow{1}{*}{Params}} & \multicolumn{3}{c|}{NJUD \cite{ju2014njud}} & \multicolumn{3}{c|}{NLPR\cite{peng2014nlpr}} & \multicolumn{3}{c|}{DUTLF-Depth\cite{Piao2019dmra}} & \multicolumn{3}{c|}{ReDWeb-S\cite{liu2021learning}} & \multicolumn{3}{c|}{STERE\cite{niu2012stere}} & \multicolumn{3}{c}{SIP\cite{fan2020SIP}} \\
  {} &{(M)} & $S_m\uparrow$ & $F_m\uparrow$ & $E_m\uparrow$ & $S_m\uparrow$ & $F_m\uparrow$ & $E_m\uparrow$ & $S_m\uparrow$ & $F_m\uparrow$ & $E_m\uparrow$ & $S_m\uparrow$ & $F_m\uparrow$ & $E_m\uparrow$ & $S_m\uparrow$ & $F_m\uparrow$ & $E_m\uparrow$ & $S_m\uparrow$ & $F_m\uparrow$ & $E_m\uparrow$\\
   \hline
   %ATST\cite{zhang2020ATSA} &32.17 &0.885 &0.893 &0.930 &0.909 &0.898 &0.951 &0.916 &0.928 &0.953 &0.679 &0.673 &0.758 &0.896 &0.901 &0.942 &0.849 &0.861 &0.901 \\
   %CMW\cite{Li2020CMWNet} &85.56 &0.870 &0.871 &0.927 &0.917 &0.903 &0.951 &0.797 &0.779 &0.864 &0.634 &0.607 &0.714 &0.852 &0.837 &0.907 &0.705 &0.677 &0.804 \\
   %Cas-Gnn\cite{luo2020Cas-Gnn} &- &0.911 &0.916 &0.948 &0.919 &0.906 &0.955 &0.920 &0.926 &0.953 & - &- &-  &0.899 &0.901 &0.944 &- &- &-   \\
   %HDFNet\cite{HDFNet-ECCV2020} &44.15 &0.908 &0.911 &0.944 &0.923 &0.917 &0.963 &0.908 &0.915 &0.945 &0.728 &0.717 &0.804 &0.900 &0.900 &0.943 &0.886 &0.894 &0.930\\
   %CoNet\cite{Wei2020CoNet} &43.66 &0.896 &0.893 &0.937 &0.912 &0.893 &0.948 &0.923 &0.932 &0.959 &0.696 &0.693 &0.782 &0.905 &0.901 &0.947 &0.860 &0.873 &0.917 \\
   %BBS-Net\cite{fan2020bbsnet} &49.77 &0.921 &0.919 &0.949 &0.931 &0.918 &0.961  &0.882 &0.870 &0.912 &0.693 &0.680 &0.763 &0.908 &0.903 &0.942 &0.879 &0.884 &0.922\\
   %JL-DCF\cite{Fu2020JLDCF} &143.52 &0.877 &0.892 &0.941 &0.931 &0.918 &\blu{0.965} &0.894 &0.891 &0.927 &0.581 &0.546 &0.708 &0.900 &0.895 &0.942 &0.885 &0.894 &0.931 \\
   %SPNet\cite{zhouiccv2021} &67.88 &0.925 &0.928 &0.957 &0.927 &0.919 &0.962 &0.895 &0.899 &0.933 &0.710 &0.715 &0.798 &0.907 &0.906 &0.949 &0.894 &0.904 &0.933 \\
   CMINet\cite{cascaded_rgbd_sod} &188.12 &{.929} &{.934} &.957 &.932 &{.922} &.963 &.912 &.913 &.938 &.725 &.726 &.800 &{.918} &{.916} &{.951} &.899 &.910 &.939\\
   %DCF\cite{Ji_2021_DCF} &53.92 &ResNet50 &0.904 &0.905 &0.943 &0.922 &0.910 &0.957 &0.925 &0.930 &0.956 &0.709 &0.715 &0.790 &0.906 &0.904 &0.948 &0.874 &0.886 &0.922 \\
   VST\cite{Liu_2021_ICCV} &53.83 &.922 &.920 &.951 &.932 &.920 &.962  &.943 &{.948} &{.969} &{.759} &{.763} &{.826}  &.913 &.907 &{.951} &{.904} &{.915} &{.944} \\
   VST-T\texttt{++} \cite{liu2023vst++} &100.51 &.928 &.929 &{.958} &{.933} &.921 &.964 &{.944} &{.948} &{.969} &.756 &.757 &.819 &.916 &.911 &.950 &.903 &.914 &{.944} \\
   SPSN\cite{lee2022spsn} &- &- &- &- &.923 &.912 &.960  &- &- &- &- &- &- &.907 &.902 &.945 &.892 &.900 &.936 \\
   CAVER\cite{pang2023caver} &55.79 &.920 &.924 &.953  &.929 &.921 &.964  &.931 &.939 &.962 &.730 &.724 &.802  &.914 &.911 &{.951} &.893 &.906 &.934 \\
    \rowcolor{c1!50}\textbf{VSCode-T} &{54.09} &\textbf{.941} &\textbf{.945} &\textbf{.967} &\textbf{.938} &\textbf{.930} &\textbf{.966} &\textbf{.952} &\textbf{.959} &\textbf{.974} &\textbf{.766} &\textbf{.771} &\textbf{.831} &\textbf{.928} &\textbf{.926} &\textbf{.957} &\textbf{.917} &\textbf{.936} &\textbf{.955} \\\hline
   \bes{VSCode-S} &{74.72} &\bes{.944} &\bes{.949} &\bes{.970} &\bes{.941} &\bes{.932} &\bes{.968} &\bes{.960} &\bes{.967} &\bes{.980}  &\bes{.777} &\bes{.776} &\bes{.829} &\bes{.931} &\bes{.928} &\bes{.958} &\bes{.924} &\bes{.942} &\bes{.958} \\
   \bottomrule[1pt]
  \end{tabular}
\vspace{-3.5mm}
  \caption{\textbf{Quantitative comparison of our VSCode with other 5 SOTA RGB-D SOD methods on six benchmark datasets.}}
  \label{RGBD_SOTA}
  \vspace{-3mm}
\end{table*}

%%%%%%%%% RGBT SOTA
\begin{table}[t]
  \centering
  \footnotesize
  %\small
  \renewcommand{\arraystretch}{1.0}
  \renewcommand{\tabcolsep}{0.6mm}
\vspace{-2mm}
%\resizebox{\linewidth}{!}{
  \begin{tabular}{l|c|ccc|ccc|ccc}
  \toprule[1pt]
    \multicolumn{1}{c|}{\multirow{2}{*}{Method}} & \multicolumn{1}{c|}{\multirow{1}{*}{Params}}&  \multicolumn{3}{c|}{VT821\cite{wang2018rgb}} & \multicolumn{3}{c|}{VT1000\cite{tu2019rgb}} & \multicolumn{3}{c}{VT5000\cite{tu2022rgbt}}\\ 
    {} &{(M)}
    & $S_m$ & $F_m$ & $E_m$ & $S_m$ & $F_m$ & $E_m$ & $S_m$ & $F_m$ & $E_m$\\\hline
    %SGDL\cite{tu2019rgb} &-  &0.765 &0.735 &0.839 &0.787 &0.770 &0.858 &0.750 &0.695 &0.829\\
    %FCMF\cite{zhang2019rgb}&- &0.760 &0.667 &0.810 &0.873 &0.851 &0.921 &0.814 &0.758 &0.866\\
    %ADF\cite{tu2022rgbt} &- &0.808 &0.749 &0.841 &0.909 &0.908 &0.950 &0.863 &0.837 &0.911\\
    %ECFFNet\cite{zhou2021ecffnet} &- &0.877 &0.835 &0.911 &0.924 &0.919 &0.959 &0.876 &0.850 &0.922\\
    %CGFNet\cite{wang2021cgfnet} &69.92 &0.881 &0.866 &0.920 &0.923 &0.923 &0.959  &0.883 &0.852 &0.926\\
    %CSRNet\cite{huo2021efficient} &1.01 &0.885 &0.855 &0.920 &0.919 &0.901 &0.952 &0.868 &0.821 &0.912\\
    %MGAI\cite{song2022multiple} &87.09 &0.891 &0.870 &0.933 &0.929 &0.921 &0.965 &0.884 &0.846 &{0.930} \\
    MIDD\cite{tu2021multi} 
    &52.43 &.871 &.847 &.916 &.916 &.904 &.956 &.868 &.834 &.919\\
    TNet\cite{cong2022does} 
    &87.41 &{.899} &{.885} &{.936} &.929 &.921 &.965 &.895 &.864 &.936 \\
    CGMDR\cite{chen2022cgmdrnet}
    &- &.894 &.872 &.932 &.931 &.927 &.966 &{.896} &{.877} &{.939}\\
    VST-T\texttt{++} \cite{liu2023vst++} 
    &100.51 &.894 &.861 &.923 &{.941} &{.931} &{.972} &.895 &.854 &.933\\
    CAVER\cite{pang2023caver} 
    &55.79 &.891 &.874 &.933 &.936 &.927 &.970 &.892 &.857 &.935
    \\
    \rowcolor{c1!50} \textbf{VSCode-T} &{54.09} &\textbf{.921} &\textbf{.906} &\textbf{.951} &\textbf{.949} &\textbf{.944} &\textbf{.981} &\textbf{.918} &\textbf{.892} &\textbf{.954} \\\hline  
    \bes{VSCode-S} &{74.72} &\bes{.926} &\bes{.910} &\bes{.954} &\bes{.952} &\bes{.947} &\bes{.981} &\bes{.925} &\bes{.900} &\bes{.959} \\
    \bottomrule[1pt]
  % \bottomrule
  \end{tabular}
  \vspace{-3.5mm}
  \caption{\textbf{Quantitative comparison of our VSCode with other 5 SOTA RGB-T SOD methods on three benchmark datasets.}}
  \label{RGBT_SOTA}
    \vspace{-4.5mm}
\end{table}

%%%%%%%%% VSOD SOTA
\begin{table*}[t]
  \centering
  %\scriptsize
  \footnotesize
  %\small
  \renewcommand{\arraystretch}{1.0}
  \renewcommand{\tabcolsep}{0.5mm}
\vspace{-2mm}
  %\resizebox{\linewidth}{!}{
  \begin{tabular}{l|c|ccc|ccc|ccc|ccc|ccc|ccc}
  \toprule[1pt]
  \multicolumn{1}{c|}{\multirow{2}{*}{Method}} & \multicolumn{1}{c|}{\multirow{1}{*}{Params}}  & \multicolumn{3}{c|}{DAVIS \cite{perazzi2016benchmark}} & \multicolumn{3}{c|}{FBMS\cite{ochs2013segmentation}} & \multicolumn{3}{c|}{ViSal\cite{wang2015consistent}} & \multicolumn{3}{c|}{SegV2\cite{li2013video}} & \multicolumn{3}{c|}{DAVSOD-Easy\cite{fan2019shifting}} & \multicolumn{3}{c}{DAVSOD-Normal\cite{fan2019shifting}}\\ 
 {} &{(M)} & $S_m\uparrow$ & $F_m\uparrow$ & $E_m\uparrow$  & $S_m\uparrow$ & $F_m\uparrow$ & $E_m\uparrow$  & $S_m\uparrow$ & $F_m\uparrow$ & $E_m\uparrow$  & $S_m\uparrow$ & $F_m\uparrow$ & $E_m\uparrow$  & $S_m\uparrow$ & $F_m\uparrow$ & $E_m\uparrow$  & $S_m\uparrow$ & $F_m\uparrow$ & $E_m\uparrow$\\
   \hline
   %SCOM\cite{chen2018scom} &- &0.820 &0.754 &0.885 &0.783 &0.778 &0.855 &0.774 &0.853 &0.938 &0.825 &0.781 &0.904 &0.572 &0.413 &0.656 &- &- &-  &- &- &-\\
   %MBN\cite{li2018unsupervised} &- &0.887 &0.862 &0.965 &0.872 &0.844 &0.910 &0.906 &0.891 &0.950 &0.763 &0.649 &0.850 &0.644 &0.479 &0.691  &- &- &- &- &- &-\\
   %PDB\cite{song2018pyramid} & &0.880 &0.851 &0.949 &0.850 &0.821 &0.882 &0.926 &0.922 &0.970 &0.871 &0.820 &0.867 &- &- &- &- &- &- &- &-\\
  % FGRN\cite{li2018flow} &- &0.839 &0.786 &0.918  &0.822 &0.783 &0.871 &0.867 &0.852 &0.954  &0.737 &0.660 &0.904 &-  &- &- &- &- &- &- &-\\
  % RCRNet\cite{yan2019semi} &53.79 &0.884 &0.845 &0.947 &0.873 &0.850 &0.902 &0.933 &0.925 &0.971 &0.829 &0.747 &0.901 &0.726 &0.601 &0.773 &0.692 &0.550 &0.760 &0.656 &0.471 &0.764\\
   %SSAV\cite{fan2019shifting} &- &0.891 &0.857 &0.945 &0.880 &0.856 &\blu{0.922} &0.944 &0.940 &0.983 &0.934 &0.797 &0.922 &- &-  &- &- &- &-  &- &-\\
   %PCSA\cite{gu2020pyramid} & &0.901 &0.878 &0.961 &0.874 &0.847 &0.914 &\blu{0.946} &0.941 &\blu{0.984} &0.887 &0.850 &0.940 &0.725 &0.590 &0.759 &- &- &- &- &- &-\\
   DCFNet\cite{zhang2021dynamic} &71.66 &.914 &.899 &.970 &{.883} &.853 &.910  &\textbf{.952} &{.953} &\textbf{.990}  &{.903} &{.870} &{.953} &.729 &.612 &.781 &.708 &.601 &.791 \\
   FSNet\cite{ji2023full} &102.30  &.922 &{.909} &{.972} &.875 &.867 &{.918}  &- &- &- &.849 &.773 &.920 &.760 &.637 &.796  &{.732} &{.623} &{.789} \\
   CoSTFormer\cite{liu2023learning} &-  &{.923} &.906 &\textbf{.978}  &- &- &- &- &- &-  &.874 &.813 &.943  &{.779} &{.667} &{.819}  &.730 &.614 &.777 \\
   UFO\cite{su2023unified} &55.92  &.918 &.906 &\textbf{.978} &.858 &{.868} &.911 &{.926} &.917 &.969  &.888 &.850 &.951  &.747 &.626 &.799  &.711 &.605 &.773\\
   \rowcolor{c1!50} \textbf{VSCode-T} &{54.09} &\textbf{.930} &\textbf{.913} &.970 &\textbf{.891} &\textbf{.880} &\textbf{.923} &\textbf{.952} &\textbf{.954} &{.989}  &\textbf{.943} &\textbf{.937} &\textbf{.984} &\textbf{.792} &\textbf{.696} &\textbf{.831}  &\textbf{.738} &\textbf{.631} &\textbf{.797} \\ \hline
   \bes{VSCode-S} &{74.72} &\bes{.936} &\bes{.922} &\bes{.973} &\bes{.905} &\bes{.902} &\bes{.939} &\bes{.955} &\bes{.957} &\bes{.991}  &\bes{.946} &\bes{.937} &\bes{.984}  &\bes{.800} &\bes{.710} &\bes{.835}  &\bes{.758} &\bes{.666} &\bes{.815} \\
   \bottomrule[1pt]
  \end{tabular}
  \vspace{-3.5mm}
  \caption{\textbf{Quantitative comparison of our VSCode with other 4 SOTA VSOD methods on six benchmark datasets.}}
  \label{VSOD_SOTA}
  \vspace{-3mm}
\end{table*}

%%%%%%%%% RGB COD SOTA
\begin{table}[t]
  \centering
  %\scriptsize
  \footnotesize
  %\small
  \renewcommand{\arraystretch}{1.0}
  \renewcommand{\tabcolsep}{0.7mm}
\vspace{-2mm}
%\resizebox{\linewidth}{!}{
  \begin{tabular}{l|c|ccc|ccc|ccc}
  \toprule[1pt]
    \multicolumn{1}{c|}{\multirow{2}{*}{Method}} & \multicolumn{1}{c|}{\multirow{1}{*}{Params}} &  \multicolumn{3}{c|}{COD10K\cite{fan2020camouflaged}} & \multicolumn{3}{c|}{NC4K\cite{lv2021simultaneously}} & \multicolumn{3}{c}{CAMO\cite{le2019anabranch}}\\
    {} &{(M)}
    & $S_m$ & $F_m$ & $E_m$ & $S_m$ & $F_m$ & $E_m$ & $S_m$ & $F_m$ & $E_m$\\\hline
    %SINet\cite{fan2020camouflaged} &48.95 &0.771 &0.676 &0.868 &0.808 &0.775 &0.883 &0.752 &0.706 &0.831 \\
    %CRLS\cite{lv2021simultaneously} &- &0.805 &0.732 &0.892 &0.840 &0.815 &0.907 &0.787 &0.753 &0.854\\
    MGL\cite{zhai2021mutual} &63.60 &.814 &.738 &.890  &- &- &- &.776 &.741 &.842 \\
    UJSC\cite{li2021uncertainty} &121.63 &.817 &.750 &.902 &{.856} &{.835} &{.920} &.803 &.775 &.867 \\
    %ZoomNet\cite{pang2022zoom} & &\blu{0.839} &\blu{0.780} &\blu{0.911} &0.853 &0.828 &0.912 &\blu{0.820} &\blu{0.805} &\blu{0.892}\\ 多输入，我们不和多输入的比较
    %DCOFD\cite{zhong2022detecting} &\blu{0.840} &\blu{0.788} &\red{0.934} &0.030 &0.834 &0.804 &0.906 &0.052 &0.844 &0.828 &0.910 &0.062\\ image_size不匹配
    SegMar\cite{jia2022segment} &56.21 &{.833} &.755 &{.907} &.841 &.827 &.907 &{.816} &{.803} &{.884} \\
    FEDER\cite{he2023camouflaged} &44.13 &.822 &{.768} &.905 &.847 &.833 &.915 &.802 &.789 &.873\\
    \rowcolor{c1!50} \textbf{VSCode-T} &{54.09} &\textbf{.847} &\textbf{.795} &\textbf{.925} &\textbf{.874} &\textbf{.853} &\textbf{.930} &\textbf{.838} &\textbf{.821} &\textbf{.909} \\\hline
    EVP\cite{liu2023explicit} &64.52 &.845 &.794 &.926 &.874 &.855 &.933 &.849 &.833 &.918 \\
    \rowcolor{c1!50} \bes{VSCode-S} &{74.72} &\bes{.869} &\bes{.827} &\bes{.942}  &\bes{.891} &\bes{.878} &\bes{.944}  &\bes{.873} &\bes{.861} &\bes{.938} \\
    \bottomrule[1pt]
  % \bottomrule
  \end{tabular}
  \vspace{-3.5mm}
  \caption{\textbf{Quantitative comparison of our VSCode with other 5 SOTA RGB COD methods on three benchmark datasets.}}
  \label{CODRGB_SOTA}
\end{table}

%%%%%%%%% VCOD SOTA
\begin{table}[h]
\centering
  \footnotesize
  \renewcommand{\arraystretch}{1.0}
  \renewcommand{\tabcolsep}{1.1mm}
  \captionsetup{justification=justified}
\vspace{-2mm}
%\begin{adjustbox}{width=\linewidth}
  \begin{tabular}{l|c|ccc|ccc}
  \toprule[1pt]
    \multicolumn{1}{c|}{\multirow{2}{*}{Method}} & \multicolumn{1}{c|}{\multirow{1}{*}{Params}} & \multicolumn{3}{c|}{CAD \cite{bideau2016s}} & \multicolumn{3}{c}{MoCA-Mask\cite{cheng2022implicit}}\\ 
   {} &{(M)} & $S_m\uparrow$ & $F_m\uparrow$ & $E_m\uparrow$ & $S_m\uparrow$ & $F_m\uparrow$ & $E_m\uparrow$ \\\hline
    PNS-Net\cite{ji2021progressively} &26.87 &.671 &.473 &.787 &.514 &.068 &.599\\
    RCRNet\cite{yan2019semi} &53.79 &.664 &.405 &.786 &.559 &.170 &.593 \\
    MG\cite{yang2021self} &- &.608 &.378 &.673 &.500 &.138 &.514\\
    SLT-Net\cite{cheng2022implicit} &164.68 &{.715} &{.542} &\textbf{.823}  &{.624} &{.327} &{.768}\\ %short-term和long-term的params之和
    \rowcolor{c1!50} \textbf{VSCode-T} &{54.09} &\textbf{.757} &\textbf{.659} &{.808} &\textbf{.650} &\textbf{.339} &\textbf{.787}\\\hline
    \bes{VSCode-S} &{74.72} &\bes{.790} &\bes{.680} &\bes{.853} &\bes{.665} &\bes{.386} &\bes{.796}\\
    \bottomrule[1pt]
  % \bottomrule
  \end{tabular}
  \vspace{-3.5mm}
  \caption{\textbf{Quantitative comparison of our VSCode with other 4 SOTA VCOD methods on two benchmark datasets.}}
  \label{CODRGBV_SOTA}
  \vspace{-4.5mm}
\end{table}
 
\vspace{-1mm}
\subsection{Ablation Study}
\vspace{-1mm}
\paragraph{Architecture Design.}
To demonstrate the efficacy of various components in our VSCode model, we report the quantitative results in Table~\ref{ALL_ablationstudy}. 
We start by performing special training (ST) on each task individually and then conduct general training (GT) on all SOD tasks. Please note that here we do not consider COD tasks since no task prompt is used. We observe improved performance on RGB-T SOD and VSOD, demonstrating the significant benefit of shared knowledge in different tasks, especially for those with limited training data diversity. 
However, the results of RGB SOD and RGB-D SOD do not show a significant increase. Our hypothesis is that amalgamating the training of multimodal images within a shared model might prevent further optimization on those well-learned tasks. 
Based on this, we introduce domain-specific prompts $p^d$, resulting in substantial improvements across all datasets, which demonstrates the efficacy of domain-specific prompts in consolidating peculiarities within their respective domains.
Subsequently, we introduce task-specific prompts $p^t$ in the encoder-decoder architecture, enabling the capability to handle COD tasks. This brings slightly improved performance on some SOD tasks, however, significantly improves the performance on all COD tasks compared with the ST baseline, which probably owes to the well-learned commonalities from different tasks.
Moreover, the incorporation of the prompt discrimination loss $\mathcal{L}_{dis}$ leads to improved performance on most tasks, reaffirming its effectiveness in disentangling peculiarities.

To further evaluate the effectiveness of the task-specific prompts in the encoder and decoder, we remove them individually, resulting in performance decrease.
This indicates that using task prompts in both encoder and decoder is necessary.
We also observe our 2D prompts only bring around 0.03M parameters, which makes our model much more parameter-efficient than the traditional special training scheme.
\footnotetext[2]{
Please note that our model shares parameters across six tasks, in contrast to EVP, which uses task-specific training. Therefore, comparing the parameters of our model with EVP may not be completely fair owing to the differences in training strategies and backbone utilization.} 

\vspace{-5mm}
\paragraph{Prompt Location.}
Following VPT \cite{jia2022visual}, we design other forms of prompt layout based on Section~\ref{sec:discus}.
Table~\ref{eachpartablationstudy} reveals that employing the shallow version of task-specific prompts in the decoder, the deep version of domain-specific prompts and task-specific prompts in the encoder yields the best results.
One plausible rationale is that each block aggregates distinct-level features within the encoder, thus it is better to propose unique prompts for each block. 
In our decoder, we follow VST and used skip connection to fuse decoder features with encoder features, which have already utilized deep task prompts for distinction. Hence, using more task prompts in the decoder may not be essential, and the shallow version seems to be a more fitting choice.

\vspace{-4mm}
\paragraph{Prompt Length.}
We perform experiments with varying lengths for three kinds of prompts.
As shown in Table~\ref{eachpartablationstudy}, 
for domain-specific prompts, using one prompt token at each block achieves better performance than using more tokens.
This suggests that it's possible to effectively capture domain distinctions using only a small number of prompts, which matches the observed relatively large correlation within domain prompts in Figure~\ref{relation}.
Regarding task-specific prompts within the encoder, a prompt layout of 1,1,5,10 tokens at four blocks is found to be optimal on COD tasks, highlighting the importance of high-level semantic features over low-level features in distinguishing between SOD and COD tasks. 
This observation matches Figure~\ref{relation} as well in which the correlations of SC in deep blocks are smaller than those in shallow blocks.
Regarding the number of task-specific prompts in the decoder, performance starts to decline when it exceeds 10. This emphasizes that blindly increasing the number of prompts doesn't guarantee improved performance. 

\vspace{-2mm}
\subsection{Comparison with State-of-the-Art Methods}
\vspace{-1mm}
Due to space limitation, we only report the performance comparison of our methods against other most highly-performed state-of-the-art methods, including 4 specialist RGB SOD models \cite{Liu_2021_ICCV,zhuge2021salient,wang2023pixels,liu2023vst++}, 5 specialist RGB-D SOD models \cite{cascaded_rgbd_sod,Liu_2021_ICCV,lee2022spsn,pang2023caver,liu2023vst++}, 5 specialist RGB-T SOD models \cite{tu2021multi,cong2022does,chen2022cgmdrnet,pang2023caver,liu2023vst++}, 3 specialist VSOD models \cite{zhang2021dynamic,ji2023full,liu2023learning}, 4 specialist RGB COD models \cite{zhai2021mutual,li2021uncertainty,jia2022segment,he2023camouflaged}, and 4 specialist VCOD models \cite{ji2021progressively,yan2019semi,yang2021self,cheng2022implicit}. Two generalist models \cite{liu2023explicit,su2023unified} are also reported.
To ensure a relatively fair comparison with EVP \cite{liu2023explicit}, which utilizes SegFormer-B4 \cite{xie2021segformer} as their backbone (64.1M parameters), we switch our backbone to Swin-S \cite{liu2021Swin} as it has a similar number of parameters (50M).
As shown in Table~\ref{RGB_SOTA}, Table~\ref{RGBD_SOTA}, Table~\ref{RGBT_SOTA}, Table~\ref{VSOD_SOTA}, Table~\ref{CODRGB_SOTA}, and Table~\ref{CODRGBV_SOTA}, our VSCode significantly outperforms all specialist methods and two generalist models across all six tasks, underscoring the effectiveness of our specially designed 2D prompts and prompt discrimination loss. 
The supplementary material displays visual comparison results among the top-performing models. 

\vspace{-1mm}
\subsection{Analysis of Generalization Ability}
\vspace{-1mm}
Previous generalist research \cite{liu2023explicit,su2023unified} primarily concentrated on assessing the effectiveness of models in training tasks, neglecting their capacity for generalizing to novel tasks. Therefore, we employ the RGB-D COD task, which is not used in training, to further investigate the zero-shot generalization capabilities of our model. 
Specifically, we utilize our well-trained model and combine depth and COD prompts to tackle the RGB-D COD task.
As shown in Table~\ref{CODRGBD_SOTA}, our VSCode model significantly outperforms the state-of-the-art specialist model PopNet \cite{wu2023source}, although ours works in a pure zero-shot way. This demonstrates the superior zero-shot generalization ability of our proposed method.
We also present the results of our model using only RGB information, which yields considerably lower performance compared to zero-shot RGB-D results. 
This validates that our zero-shot performance is not reliant on the utilization of seen RGB COD information but on the effectiveness of consolidating domain- and task-specific knowledge, which allows for the straightforward combination of various domain- and task-specific prompts for unseen tasks.

%%%%%%%%% RGB-D COD SOTA
\begin{table}[t]
  \centering
  %\scriptsize
  \footnotesize
  %\small
  \renewcommand{\arraystretch}{1.0}
  \renewcommand{\tabcolsep}{0.55mm}
\vspace{-2mm}
%\resizebox{\linewidth}{!}{
  \begin{tabular}{l|c|ccc|ccc|ccc}
  \toprule[1pt]
    \multicolumn{2}{c|}{Summary} &  \multicolumn{3}{c|}{COD10K\cite{fan2020camouflaged}} & \multicolumn{3}{c|}{NC4K\cite{lv2021simultaneously}} & \multicolumn{3}{c}{CAMO\cite{le2019anabranch}}\\ 
    \cline{0-1}
    \multicolumn{1}{c|}{Method} 
    &  \multicolumn{1}{c|}{Task} & $S_m$ & $F_m$ & $E_m$ & $S_m$ & $F_m$ & $E_m$ & $S_m$ & $F_m$ & $E_m$ \\\hline
    PopNet\cite{wu2023source} & {RGB-D} &.851 &.802 &.919 &.861 &.843 &.919 &.808 &.792 &.874 \\
    \rowcolor{c1!50} \textbf{VSCode-T} & \textbf{ZS RGB-D}&\textbf{.882} &\textbf{.849} &\textbf{.945} &\textbf{.902} &\textbf{.894} &\textbf{.950}  &\textbf{.885} &\textbf{.885} &\textbf{.945} \\\hline
    VSCode-T & {RGB} &.847 &.795 &.925 &.874 &.853 &.930 &.838 &.821 &.909\\
  \bottomrule[1pt]
  \end{tabular}
  \vspace{-3.5mm}
  \caption{\textbf{Comparison with the SOTA RGB-D COD method on three benchmark datasets.} ``ZS" indicates zero-shot.}
  \label{CODRGBD_SOTA}
  \vspace{-5.5mm}
\end{table}

\vspace{-2mm}
\section{Conclusion}
\vspace{-2mm}
In this paper, we present VSCode, a novel generalist and parameter-efficient model that tackles general multimodal SOD and COD tasks. Concretely, we use a foundation model to assimilate commonalities and 2D prompts to learn domain and task peculiarities. 
Furthermore, a prompt discrimination loss is introduced to help effectively disentangle specific knowledge and learn better shared knowledge. Our experiments demonstrate the effectiveness of VSCode on six training tasks and one unseen task. 
% 讨论加不进去了
%In the future, we will integrate the concept of 2D prompts into large multi-task models, such as X-decoder, for zero-shot multimodal object detection or segmentation. 
\vspace{-2mm}
\section*{Acknowledgments}
\vspace{-2mm}
\footnotesize
This work was supported in part by the Key R\&D Program of Shaanxi Province under Grant 2021ZDLGY01-08; the National Natural Science Foundation of China under Grants 62136007, U20B2065, 62036005, 62322605; the Key Research and Development Program of Jiangsu Province under Grant BE2021093; the Institute of Artificial Intelligence, Hefei Comprehensive National Science Center Project under Grant 21KT008; and by the MBZUAI-WIS Joint Program for AI Research under Grants WIS P008 and P009.

%%%%%%%%% REFERENCES
{\small
\bibliographystyle{ieee_fullname}
\bibliography{main}

\begin{thebibliography}{100}\itemsep=-1pt

\bibitem{awais2023foundational}
Muhammad Awais, Muzammal Naseer, Salman Khan, Rao~Muhammad Anwer, Hisham Cholakkal, Mubarak Shah, Ming-Hsuan Yang, and Fahad~Shahbaz Khan.
\newblock Foundational models defining a new era in vision: A survey and outlook.
\newblock {\em arXiv preprint arXiv:2307.13721}, 2023.

\bibitem{ba2016layer}
Jimmy~Lei Ba, Jamie~Ryan Kiros, and Geoffrey~E Hinton.
\newblock Layer normalization.
\newblock {\em arXiv preprint arXiv:1607.06450}, 2016.

\bibitem{bideau2016s}
Pia Bideau and Erik Learned-Miller.
\newblock It’s moving! a probabilistic model for causal motion segmentation in moving camera videos.
\newblock In {\em ECCV}, pages 433--449. Springer, 2016.

\bibitem{bideau2018moa}
Pia Bideau, Rakesh~R Menon, and Erik Learned-Miller.
\newblock Moa-net: self-supervised motion segmentation.
\newblock In {\em ECCV}, pages 0--0, 2018.

\bibitem{borji2019salient}
Ali Borji, Ming-Ming Cheng, Qibin Hou, Huaizu Jiang, and Jia Li.
\newblock Salient object detection: A survey.
\newblock {\em CVMJ}, 5:117--150, 2019.

\bibitem{brown2020language}
Tom Brown, Benjamin Mann, Nick Ryder, Melanie Subbiah, Jared~D Kaplan, Prafulla Dhariwal, Arvind Neelakantan, Pranav Shyam, Girish Sastry, Amanda Askell, et~al.
\newblock Language models are few-shot learners.
\newblock {\em NeurIPS}, 33:1877--1901, 2020.

\bibitem{chen2022cgmdrnet}
Gang Chen, Feng Shao, Xiongli Chai, Hangwei Chen, Qiuping Jiang, Xiangchao Meng, and Yo-Sung Ho.
\newblock Cgmdrnet: Cross-guided modality difference reduction network for rgb-t salient object detection.
\newblock {\em IEEE TCSVT}, 32(9):6308--6323, 2022.

\bibitem{chen2020PGAR}
Shuhan Chen and Yun Fu.
\newblock Progressively guided alternate refinement network for rgb-d salient object detection.
\newblock In {\em ECCV}, pages 520--538, 2020.

\bibitem{chen2020dpanet}
Zuyao Chen, Runmin Cong, Qianqian Xu, and Qingming Huang.
\newblock Dpanet: Depth potentiality-aware gated attention network for rgb-d salient object detection.
\newblock {\em IEEE TIP}, 2020.

\bibitem{cheng2022implicit}
Xuelian Cheng, Huan Xiong, Deng-Ping Fan, Yiran Zhong, Mehrtash Harandi, Tom Drummond, and Zongyuan Ge.
\newblock Implicit motion handling for video camouflaged object detection.
\newblock In {\em CVPR}, pages 13864--13873, 2022.

\bibitem{cong2022does}
Runmin Cong, Kepu Zhang, Chen Zhang, Feng Zheng, Yao Zhao, Qingming Huang, and Sam Kwong.
\newblock Does thermal really always matter for rgb-t salient object detection?
\newblock {\em IEEE TMM}, 2022.

\bibitem{deng2018r3net}
Zijun Deng, Xiaowei Hu, Lei Zhu, Xuemiao Xu, Jing Qin, Guoqiang Han, and Pheng-Ann Heng.
\newblock R3net: Recurrent residual refinement network for saliency detection.
\newblock In {\em IJCAI}, pages 684--690, 2018.

\bibitem{fan2017structure}
Deng-Ping Fan, Ming-Ming Cheng, Yun Liu, Tao Li, and Ali Borji.
\newblock Structure-measure: A new way to evaluate foreground maps.
\newblock In {\em ICCV}, pages 4548--4557, 2017.

\bibitem{Fan2018Enhanced}
Deng-Ping Fan, Cheng Gong, Yang Cao, Bo Ren, Ming-Ming Cheng, and Ali Borji.
\newblock {Enhanced-alignment Measure for Binary Foreground Map Evaluation}.
\newblock In {\em IJCAI}, pages 698--704, 2018.

\bibitem{fan2020camouflaged}
Deng-Ping Fan, Ge-Peng Ji, Guolei Sun, Ming-Ming Cheng, Jianbing Shen, and Ling Shao.
\newblock Camouflaged object detection.
\newblock In {\em CVPR}, pages 2777--2787, 2020.

\bibitem{fan2023advances}
Deng-Ping Fan, Ge-Peng Ji, Peng Xu, Ming-Ming Cheng, Christos Sakaridis, and Luc Van~Gool.
\newblock Advances in deep concealed scene understanding.
\newblock {\em VI}, 1(1):16, 2023.

\bibitem{fan2020SIP}
Deng-Ping Fan, Zheng Lin, Zhao Zhang, Menglong Zhu, and Ming-Ming Cheng.
\newblock Rethinking rgb-d salient object detection: Models, data sets, and large-scale benchmarks.
\newblock {\em IEEE TNNLS}, 32(5):2075--2089, 2020.

\bibitem{fan2019shifting}
Deng-Ping Fan, Wenguan Wang, Ming-Ming Cheng, and Jianbing Shen.
\newblock Shifting more attention to video salient object detection.
\newblock In {\em CVPR}, pages 8554--8564, 2019.

\bibitem{fan2020bbsnet}
Deng-Ping Fan, Yingjie Zhai, Ali Borji, Jufeng Yang, and Ling Shao.
\newblock Bbs-net: Rgb-d salient object detection with a bifurcated backbone strategy network.
\newblock In {\em ECCV}, pages 275--292, 2020.

\bibitem{fang2022densely}
Chaowei Fang, Haibin Tian, Dingwen Zhang, Qiang Zhang, Jungong Han, and Junwei Han.
\newblock Densely nested top-down flows for salient object detection.
\newblock {\em SCIS}, 65(8):182103, 2022.

\bibitem{gu2020pyramid}
Yuchao Gu, Lijuan Wang, Ziqin Wang, Yun Liu, Ming-Ming Cheng, and Shao-Ping Lu.
\newblock Pyramid constrained self-attention network for fast video salient object detection.
\newblock In {\em AAAI}, volume~34, pages 10869--10876, 2020.

\bibitem{he2023camouflaged}
Chunming He, Kai Li, Yachao Zhang, Longxiang Tang, Yulun Zhang, Zhenhua Guo, and Xiu Li.
\newblock Camouflaged object detection with feature decomposition and edge reconstruction.
\newblock In {\em CVPR}, pages 22046--22055, 2023.

\bibitem{he2016resnet}
Kaiming He, Xiangyu Zhang, Shaoqing Ren, and Jian Sun.
\newblock Deep residual learning for image recognition.
\newblock In {\em CVPR}, pages 770--778, 2016.

\bibitem{hou2018dss}
Q Hou, MM Cheng, X Hu, A Borji, Z Tu, and PHS Torr.
\newblock Deeply supervised salient object detection with short connections.
\newblock {\em IEEE TPAMI}, 41(4):815--828, 2018.

\bibitem{hu2023compositional}
Guyue Hu, Bin He, and Hanwang Zhang.
\newblock Compositional prompting video-language models to understand procedure in instructional videos.
\newblock {\em MIR}, 20(2):249--262, 2023.

\bibitem{ji2021progressively}
Ge-Peng Ji, Yu-Cheng Chou, Deng-Ping Fan, Geng Chen, Huazhu Fu, Debesh Jha, and Ling Shao.
\newblock Progressively normalized self-attention network for video polyp segmentation.
\newblock In {\em MICCAI}, pages 142--152. Springer, 2021.

\bibitem{ji2023deep}
Ge-Peng Ji, Deng-Ping Fan, Yu-Cheng Chou, Dengxin Dai, Alexander Liniger, and Luc Van~Gool.
\newblock Deep gradient learning for efficient camouflaged object detection.
\newblock {\em MIR}, 20(1):92--108, 2023.

\bibitem{ji2023full}
Ge-Peng Ji, Deng-Ping Fan, Keren Fu, Zhe Wu, Jianbing Shen, and Ling Shao.
\newblock Full-duplex strategy for video object segmentation.
\newblock In {\em CVMJ}, pages 155--175, 2023.

\bibitem{Wei2020CoNet}
Wei Ji, Jingjing Li, Miao Zhang, Yongri Piao, and Huchuan Lu.
\newblock Accurate rgb-d salient object detection via collaborative learning.
\newblock In {\em ECCV}, pages 52--69, 2020.

\bibitem{ji2020casnet}
Yuzhu Ji, Haijun Zhang, Zequn Jie, Lin Ma, and QM~Jonathan Wu.
\newblock Casnet: A cross-attention siamese network for video salient object detection.
\newblock {\em IEEE TNNLS}, 32(6):2676--2690, 2020.

\bibitem{jia2022visual}
Menglin Jia, Luming Tang, Bor-Chun Chen, Claire Cardie, Serge Belongie, Bharath Hariharan, and Ser-Nam Lim.
\newblock Visual prompt tuning.
\newblock In {\em ECCV}, pages 709--727. Springer, 2022.

\bibitem{jia2022segment}
Qi Jia, Shuilian Yao, Yu Liu, Xin Fan, Risheng Liu, and Zhongxuan Luo.
\newblock Segment, magnify and reiterate: Detecting camouflaged objects the hard way.
\newblock In {\em CVPR}, pages 4713--4722, 2022.

\bibitem{ju2014njud}
Ran Ju, Ling Ge, Wenjing Geng, Tongwei Ren, and Gangshan Wu.
\newblock Depth saliency based on anisotropic center-surround difference.
\newblock In {\em ICIP}, pages 1115--1119, 2014.

\bibitem{kendall2018multi}
Alex Kendall, Yarin Gal, and Roberto Cipolla.
\newblock Multi-task learning using uncertainty to weigh losses for scene geometry and semantics.
\newblock In {\em CVPR}, pages 7482--7491, 2018.

\bibitem{kingma2014adam}
Diederik~P Kingma and Jimmy Ba.
\newblock Adam: A method for stochastic optimization.
\newblock {\em arXiv preprint arXiv:1412.6980}, 2014.

\bibitem{lamdouar2020betrayed}
Hala Lamdouar, Charig Yang, Weidi Xie, and Andrew Zisserman.
\newblock Betrayed by motion: Camouflaged object discovery via motion segmentation.
\newblock In {\em ACCV}, 2020.

\bibitem{le2019anabranch}
Trung-Nghia Le, Tam~V Nguyen, Zhongliang Nie, Minh-Triet Tran, and Akihiro Sugimoto.
\newblock Anabranch network for camouflaged object segmentation.
\newblock {\em CVIU}, 184:45--56, 2019.

\bibitem{lee2022spsn}
Minhyeok Lee, Chaewon Park, Suhwan Cho, and Sangyoun Lee.
\newblock Spsn: Superpixel prototype sampling network for rgb-d salient object detection.
\newblock In {\em ECCV}, pages 630--647. Springer, 2022.

\bibitem{li2021uncertainty}
Aixuan Li, Jing Zhang, Yunqiu Lv, Bowen Liu, Tong Zhang, and Yuchao Dai.
\newblock Uncertainty-aware joint salient object and camouflaged object detection.
\newblock In {\em CVPR}, pages 10071--10081, 2021.

\bibitem{li2023joint}
Aixuan Li, Jing Zhang, Yunqiu Lv, Tong Zhang, Yiran Zhong, Mingyi He, and Yuchao Dai.
\newblock Joint salient object detection and camouflaged object detection via uncertainty-aware learning.
\newblock {\em arXiv preprint arXiv:2307.04651}, 2023.

\bibitem{li2013video}
Fuxin Li, Taeyoung Kim, Ahmad Humayun, David Tsai, and James~M Rehg.
\newblock Video segmentation by tracking many figure-ground segments.
\newblock In {\em ICCV}, pages 2192--2199, 2013.

\bibitem{li2020icnet}
Gongyang Li, Zhi Liu, and Haibin Ling.
\newblock Icnet: Information conversion network for rgb-d based salient object detection.
\newblock {\em IEEE TIP}, 29:4873--4884, 2020.

\bibitem{Li2020CMWNet}
Gongyang Li, Zhi Liu, Linwei Ye, Yang Wang, and Haibin Ling.
\newblock Cross-modal weighting network for rgb-d salient object detection.
\newblock In {\em ECCV}, pages 665--681, 2020.

\bibitem{li2018flow}
Guanbin Li, Yuan Xie, Tianhao Wei, Keze Wang, and Liang Lin.
\newblock Flow guided recurrent neural encoder for video salient object detection.
\newblock In {\em CVPR}, pages 3243--3252, 2018.

\bibitem{li2015HKUIS}
Guanbin Li and Yizhou Yu.
\newblock Visual saliency based on multiscale deep features.
\newblock In {\em CVPR}, pages 5455--5463, 2015.

\bibitem{li2023boosting}
Hao Li, Dingwen Zhang, Nian Liu, Lechao Cheng, Yalun Dai, Chao Zhang, Xinggang Wang, and Junwei Han.
\newblock Boosting low-data instance segmentation by unsupervised pre-training with saliency prompt.
\newblock In {\em CVPR}, pages 15485--15494, 2023.

\bibitem{li2014PASCALS}
Yin Li, Xiaodi Hou, Christof Koch, James~M Rehg, and Alan~L Yuille.
\newblock The secrets of salient object segmentation.
\newblock In {\em CVPR}, pages 280--287, 2014.

\bibitem{liu2016dhsnet}
Nian Liu and Junwei Han.
\newblock Dhsnet: Deep hierarchical saliency network for salient object detection.
\newblock In {\em CVPR}, pages 678--686, 2016.

\bibitem{liu2018picanet}
Nian Liu, Junwei Han, and Ming-Hsuan Yang.
\newblock Picanet: Learning pixel-wise contextual attention for saliency detection.
\newblock In {\em CVPR}, pages 3089--3098, 2018.

\bibitem{liu2023vst++}
Nian Liu, Ziyang Luo, Ni Zhang, and Junwei Han.
\newblock Vst++: Efficient and stronger visual saliency transformer.
\newblock {\em arXiv preprint arXiv:2310.11725}, 2023.

\bibitem{liu2023learning}
Nian Liu, Kepan Nan, Wangbo Zhao, Xiwen Yao, and Junwei Han.
\newblock Learning complementary spatial--temporal transformer for video salient object detection.
\newblock {\em IEEE TNNLS}, 2023.

\bibitem{liu2020S2MA}
Nian Liu, Ni Zhang, and Junwei Han.
\newblock Learning selective self-mutual attention for rgb-d saliency detection.
\newblock In {\em CVPR}, pages 13756--13765, 2020.

\bibitem{liu2021learning}
Nian Liu, Ni Zhang, Ling Shao, and Junwei Han.
\newblock Learning selective mutual attention and contrast for rgb-d saliency detection.
\newblock {\em IEEE TPAMI}, 44(12):9026--9042, 2021.

\bibitem{Liu_2021_ICCV}
Nian Liu, Ni Zhang, Kaiyuan Wan, Ling Shao, and Junwei Han.
\newblock Visual saliency transformer.
\newblock In {\em ICCV}, pages 4722--4732, October 2021.

\bibitem{liu2019end}
Shikun Liu, Edward Johns, and Andrew~J Davison.
\newblock End-to-end multi-task learning with attention.
\newblock In {\em CVPR}, pages 1871--1880, 2019.

\bibitem{liu2023explicit}
Weihuang Liu, Xi Shen, Chi-Man Pun, and Xiaodong Cun.
\newblock Explicit visual prompting for universal foreground segmentations.
\newblock {\em arXiv preprint arXiv:2305.18476}, 2023.

\bibitem{liu2022disentangled}
Yi Liu, Dingwen Zhang, Nian Liu, Shoukun Xu, and Jungong Han.
\newblock Disentangled capsule routing for fast part-object relational saliency.
\newblock {\em IEEE TIP}, 31:6719--6732, 2022.

\bibitem{liu2021part}
Yi Liu, Dingwen Zhang, Qiang Zhang, and Jungong Han.
\newblock Part-object relational visual saliency.
\newblock {\em TPAMI}, 44(7):3688--3704, 2021.

\bibitem{liu2021Swin}
Ze Liu, Yutong Lin, Yue Cao, Han Hu, Yixuan Wei, Zheng Zhang, Stephen Lin, and Baining Guo.
\newblock Swin transformer: Hierarchical vision transformer using shifted windows.
\newblock In {\em ICCV}, 2021.

\bibitem{liu2019salient}
Zhengyi Liu, Song Shi, Quntao Duan, Wei Zhang, and Peng Zhao.
\newblock Salient object detection for rgb-d image by single stream recurrent convolution neural network.
\newblock {\em IJON}, 363:46--57, 2019.

\bibitem{lv2021simultaneously}
Yunqiu Lv, Jing Zhang, Yuchao Dai, Aixuan Li, Bowen Liu, Nick Barnes, and Deng-Ping Fan.
\newblock Simultaneously localize, segment and rank the camouflaged objects.
\newblock In {\em CVPR}, pages 11591--11601, 2021.

\bibitem{movahedi2010SOD}
Vida Movahedi and James~H Elder.
\newblock Design and perceptual validation of performance measures for salient object segmentation.
\newblock In {\em CVPR}, pages 49--56, 2010.

\bibitem{nie2022pro}
Xing Nie, Bolin Ni, Jianlong Chang, Gaomeng Meng, Chunlei Huo, Zhaoxiang Zhang, Shiming Xiang, Qi Tian, and Chunhong Pan.
\newblock Pro-tuning: Unified prompt tuning for vision tasks.
\newblock {\em arXiv preprint arXiv:2207.14381}, 2022.

\bibitem{niu2012stere}
Yuzhen Niu, Yujie Geng, Xueqing Li, and Feng Liu.
\newblock Leveraging stereopsis for saliency analysis.
\newblock In {\em CVPR}, pages 454--461, 2012.

\bibitem{ochs2013segmentation}
Peter Ochs, Jitendra Malik, and Thomas Brox.
\newblock Segmentation of moving objects by long term video analysis.
\newblock {\em IEEE TPAMI}, 36(6):1187--1200, 2013.

\bibitem{pang2022zoom}
Youwei Pang, Xiaoqi Zhao, Tian-Zhu Xiang, Lihe Zhang, and Huchuan Lu.
\newblock Zoom in and out: A mixed-scale triplet network for camouflaged object detection.
\newblock In {\em CVPR}, pages 2160--2170, 2022.

\bibitem{MINet-CVPR2020}
Youwei Pang, Xiaoqi Zhao, Lihe Zhang, and Huchuan Lu.
\newblock Multi-scale interactive network for salient object detection.
\newblock In {\em CVPR}, pages 9413--9422, 2020.

\bibitem{pang2023caver}
Youwei Pang, Xiaoqi Zhao, Lihe Zhang, and Huchuan Lu.
\newblock Caver: Cross-modal view-mixed transformer for bi-modal salient object detection.
\newblock {\em IEEE TIP}, 32:892--904, 2023.

\bibitem{peng2014nlpr}
Houwen Peng, Bing Li, Weihua Xiong, Weiming Hu, and Rongrong Ji.
\newblock Rgbd salient object detection: A benchmark and algorithms.
\newblock In {\em ECCV}, pages 92--109, 2014.

\bibitem{perazzi2016benchmark}
Federico Perazzi, Jordi Pont-Tuset, Brian McWilliams, Luc Van~Gool, Markus Gross, and Alexander Sorkine-Hornung.
\newblock A benchmark dataset and evaluation methodology for video object segmentation.
\newblock In {\em CVPR}, pages 724--732, 2016.

\bibitem{Piao2019dmra}
Yongri Piao, Wei Ji, Jingjing Li, Miao Zhang, and Huchuan Lu.
\newblock Depth-induced multi-scale recurrent attention network for saliency detection.
\newblock In {\em ICCV}, pages 7254--7263, 2019.

\bibitem{qin2019basnet}
Xuebin Qin, Zichen Zhang, Chenyang Huang, Chao Gao, Masood Dehghan, and Martin Jagersand.
\newblock Basnet: Boundary-aware salient object detection.
\newblock In {\em CVPR}, pages 7479--7489, 2019.

\bibitem{ren2020tenet}
Sucheng Ren, Chu Han, Xin Yang, Guoqiang Han, and Shengfeng He.
\newblock Tenet: Triple excitation network for video salient object detection.
\newblock In {\em ECCV}, pages 212--228. Springer, 2020.

\bibitem{song2018pyramid}
Hongmei Song, Wenguan Wang, Sanyuan Zhao, Jianbing Shen, and Kin-Man Lam.
\newblock Pyramid dilated deeper convlstm for video salient object detection.
\newblock In {\em ECCV}, pages 715--731, 2018.

\bibitem{su2023unified}
Yukun Su, Jingliang Deng, Ruizhou Sun, Guosheng Lin, Hanjing Su, and Qingyao Wu.
\newblock A unified transformer framework for group-based segmentation: Co-segmentation, co-saliency detection and video salient object detection.
\newblock {\em IEEE TMM}, 2023.

\bibitem{tu2021multi}
Zhengzheng Tu, Zhun Li, Chenglong Li, Yang Lang, and Jin Tang.
\newblock Multi-interactive dual-decoder for rgb-thermal salient object detection.
\newblock {\em IEEE TIP}, 30:5678--5691, 2021.

\bibitem{tu2022rgbt}
Zhengzheng Tu, Yan Ma, Zhun Li, Chenglong Li, Jieming Xu, and Yongtao Liu.
\newblock Rgbt salient object detection: A large-scale dataset and benchmark.
\newblock {\em IEEE TMM}, 2022.

\bibitem{tu2019rgb}
Zhengzheng Tu, Tian Xia, Chenglong Li, Xiaoxiao Wang, Yan Ma, and Jin Tang.
\newblock Rgb-t image saliency detection via collaborative graph learning.
\newblock {\em IEEE TMM}, 22(1):160--173, 2019.

\bibitem{wang2018rgb}
Guizhao Wang, Chenglong Li, Yunpeng Ma, Aihua Zheng, Jin Tang, and Bin Luo.
\newblock Rgb-t saliency detection benchmark: Dataset, baselines, analysis and a novel approach.
\newblock In {\em IJIG}, pages 359--369. Springer, 2018.

\bibitem{wang2021cgfnet}
Jie Wang, Kechen Song, Yanqi Bao, Liming Huang, and Yunhui Yan.
\newblock Cgfnet: Cross-guided fusion network for rgb-t salient object detection.
\newblock {\em IEEE TCSVT}, 32(5):2949--2961, 2021.

\bibitem{wang2017duts}
Lijun Wang, Huchuan Lu, Yifan Wang, Mengyang Feng, Dong Wang, Baocai Yin, and Xiang Ruan.
\newblock Learning to detect salient objects with image-level supervision.
\newblock In {\em CVPR}, pages 136--145, 2017.

\bibitem{wang2018rfcn}
Linzhao Wang, Lijun Wang, Huchuan Lu, Pingping Zhang, and Xiang Ruan.
\newblock Salient object detection with recurrent fully convolutional networks.
\newblock {\em IEEE TPAMI}, 41(7):1734--1746, 2018.

\bibitem{wang2017stagewise}
Tiantian Wang, Ali Borji, Lihe Zhang, Pingping Zhang, and Huchuan Lu.
\newblock A stagewise refinement model for detecting salient objects in images.
\newblock In {\em ICCV}, pages 4019--4028, 2017.

\bibitem{wang2018salient}
Wenguan Wang, Jianbing Shen, Xingping Dong, and Ali Borji.
\newblock Salient object detection driven by fixation prediction.
\newblock In {\em CVPR}, pages 1711--1720, 2018.

\bibitem{wang2015consistent}
Wenguan Wang, Jianbing Shen, and Ling Shao.
\newblock Consistent video saliency using local gradient flow optimization and global refinement.
\newblock {\em IEEE TIP}, 24(11):4185--4196, 2015.

\bibitem{wang2017video}
Wenguan Wang, Jianbing Shen, and Ling Shao.
\newblock Video salient object detection via fully convolutional networks.
\newblock {\em IEEE TIP}, 27(1):38--49, 2017.

\bibitem{wang2023images}
Xinlong Wang, Wen Wang, Yue Cao, Chunhua Shen, and Tiejun Huang.
\newblock Images speak in images: A generalist painter for in-context visual learning.
\newblock In {\em CVPR}, pages 6830--6839, 2023.

\bibitem{wang2023seggpt}
Xinlong Wang, Xiaosong Zhang, Yue Cao, Wen Wang, Chunhua Shen, and Tiejun Huang.
\newblock Seggpt: Segmenting everything in context.
\newblock {\em arXiv preprint arXiv:2304.03284}, 2023.

\bibitem{wang2023pixels}
Yi Wang, Ruili Wang, Xin Fan, Tianzhu Wang, and Xiangjian He.
\newblock Pixels, regions, and objects: Multiple enhancement for salient object detection.
\newblock In {\em CVPR}, pages 10031--10040, 2023.

\bibitem{CVPR2020_LDF}
Jun Wei, Shuhui Wang, Zhe Wu, Chi Su, Qingming Huang, and Qi Tian.
\newblock Label decoupling framework for salient object detection.
\newblock In {\em CVPR}, pages 13025--13034, 2020.

\bibitem{wei2020end}
Lina Wei, Shanshan Zhao, Omar~Farouk Bourahla, Xi Li, Fei Wu, Yueting Zhuang, Junwei Han, and Mingliang Xu.
\newblock End-to-end video saliency detection via a deep contextual spatiotemporal network.
\newblock {\em IEEE TNNLS}, 32(4):1691--1702, 2020.

\bibitem{wu2023source}
Zongwei Wu, Danda~Pani Paudel, Deng-Ping Fan, Jingjing Wang, Shuo Wang, C{\'e}dric Demonceaux, Radu Timofte, and Luc Van~Gool.
\newblock Source-free depth for object pop-out.
\newblock In {\em ICCV}, pages 1032--1042, 2023.

\bibitem{xie2021segformer}
Enze Xie, Wenhai Wang, Zhiding Yu, Anima Anandkumar, Jose~M Alvarez, and Ping Luo.
\newblock Segformer: Simple and efficient design for semantic segmentation with transformers.
\newblock {\em NeurIPS}, 34:12077--12090, 2021.

\bibitem{xie2020count}
Jin Xie, Hisham Cholakkal, Rao Muhammad~Anwer, Fahad Shahbaz~Khan, Yanwei Pang, Ling Shao, and Mubarak Shah.
\newblock Count-and similarity-aware r-cnn for pedestrian detection.
\newblock In {\em ECCV}, pages 88--104. Springer, 2020.

\bibitem{yan2023universal}
Bin Yan, Yi Jiang, Jiannan Wu, Dong Wang, Ping Luo, Zehuan Yuan, and Huchuan Lu.
\newblock Universal instance perception as object discovery and retrieval.
\newblock In {\em CVPR}, pages 15325--15336, 2023.

\bibitem{yan2019semi}
Pengxiang Yan, Guanbin Li, Yuan Xie, Zhen Li, Chuan Wang, Tianshui Chen, and Liang Lin.
\newblock Semi-supervised video salient object detection using pseudo-labels.
\newblock In {\em ICCV}, pages 7284--7293, 2019.

\bibitem{yan2013ECSSD}
Qiong Yan, Li Xu, Jianping Shi, and Jiaya Jia.
\newblock Hierarchical saliency detection.
\newblock In {\em CVPR}, pages 1155--1162, 2013.

\bibitem{yang2021self}
Charig Yang, Hala Lamdouar, Erika Lu, Andrew Zisserman, and Weidi Xie.
\newblock Self-supervised video object segmentation by motion grouping.
\newblock In {\em ICCV}, pages 7177--7188, 2021.

\bibitem{yang2013DUTO}
Chuan Yang, Lihe Zhang, Huchuan Lu, Xiang Ruan, and Ming-Hsuan Yang.
\newblock Saliency detection via graph-based manifold ranking.
\newblock In {\em CVPR}, pages 3166--3173, 2013.

\bibitem{zeiler2014visualizing}
Matthew~D Zeiler and Rob Fergus.
\newblock Visualizing and understanding convolutional networks.
\newblock In {\em ECCV}, pages 818--833. Springer, 2014.

\bibitem{zhai2021mutual}
Qiang Zhai, Xin Li, Fan Yang, Chenglizhao Chen, Hong Cheng, and Deng-Ping Fan.
\newblock Mutual graph learning for camouflaged object detection.
\newblock In {\em CVPR}, pages 12997--13007, 2021.

\bibitem{zhang2019synthesizing}
Dingwen Zhang, Junwei Han, Yu Zhang, and Dong Xu.
\newblock Synthesizing supervision for learning deep saliency network without human annotation.
\newblock {\em IEEE TPAMI}, 42(7):1755--1769, 2019.

\bibitem{cascaded_rgbd_sod}
Jing Zhang, Deng-Ping Fan, Yuchao Dai, Xin Yu, Yiran Zhong, Nick Barnes, and Ling Shao.
\newblock Rgb-d saliency detection via cascaded mutual information minimization.
\newblock In {\em ICCV}, 2021.

\bibitem{zhang2019capsal}
Lu Zhang, Jianming Zhang, Zhe Lin, Huchuan Lu, and You He.
\newblock Capsal: Leveraging captioning to boost semantics for salient object detection.
\newblock In {\em CVPR}, pages 6024--6033, 2019.

\bibitem{zhang2021dynamic}
Miao Zhang, Jie Liu, Yifei Wang, Yongri Piao, Shunyu Yao, Wei Ji, Jingjing Li, Huchuan Lu, and Zhongxuan Luo.
\newblock Dynamic context-sensitive filtering network for video salient object detection.
\newblock In {\em ICCV}, pages 1553--1563, 2021.

\bibitem{zhang2020select}
Miao Zhang, Weisong Ren, Yongri Piao, Zhengkun Rong, and Huchuan Lu.
\newblock Select, supplement and focus for rgb-d saliency detection.
\newblock In {\em CVPR}, pages 3472--3481, 2020.

\bibitem{zhang2019rgb}
Qiang Zhang, Nianchang Huang, Lin Yao, Dingwen Zhang, Caifeng Shan, and Jungong Han.
\newblock Rgb-t salient object detection via fusing multi-level cnn features.
\newblock {\em IEEE TIP}, 29:3321--3335, 2019.

\bibitem{zhang2018pagr}
Xiaoning Zhang, Tiantian Wang, Jinqing Qi, Huchuan Lu, and Gang Wang.
\newblock Progressive attention guided recurrent network for salient object detection.
\newblock In {\em CVPR}, pages 714--722, 2018.

\bibitem{zhao2019contrast}
Jia-Xing Zhao, Yang Cao, Deng-Ping Fan, Ming-Ming Cheng, Xuan-Yi Li, and Le Zhang.
\newblock Contrast prior and fluid pyramid integration for rgbd salient object detection.
\newblock In {\em CVPR}, pages 3927--3936, 2019.

\bibitem{zhao2019EGNet}
Jia-Xing Zhao, Jiang-Jiang Liu, Deng-Ping Fan, Yang Cao, Jufeng Yang, and Ming-Ming Cheng.
\newblock Egnet:edge guidance network for salient object detection.
\newblock In {\em ICCV}, pages 8779--8788, 2019.

\bibitem{zhao2023learning}
Wangbo Zhao, Kepan Nan, Songyang Zhang, Kai Chen, Dahua Lin, and Yang You.
\newblock Learning referring video object segmentation from weak annotation.
\newblock {\em arXiv preprint arXiv:2308.02162}, 2023.

\bibitem{zhao2024dynamic}
Wangbo Zhao, Jiasheng Tang, Yizeng Han, Yibing Song, Kai Wang, Gao Huang, Fan Wang, and Yang You.
\newblock Dynamic tuning towards parameter and inference efficiency for vit adaptation.
\newblock {\em arXiv preprint arXiv:2403.11808}, 2024.

\bibitem{GateNet}
Xiaoqi Zhao, Youwei Pang, Lihe Zhang, Huchuan Lu, and Lei Zhang.
\newblock Suppress and balance: A simple gated network for salient object detection.
\newblock In {\em ECCV}, pages 35--51, 2020.

\bibitem{zheng2023mffn}
Dehua Zheng, Xiaochen Zheng, Laurence~T Yang, Yuan Gao, Chenlu Zhu, and Yiheng Ruan.
\newblock Mffn: Multi-view feature fusion network for camouflaged object detection.
\newblock In {\em WACV}, pages 6232--6242, 2023.

\bibitem{zhu2023visual}
Jiawen Zhu, Simiao Lai, Xin Chen, Dong Wang, and Huchuan Lu.
\newblock Visual prompt multi-modal tracking.
\newblock In {\em CVPR}, pages 9516--9526, 2023.

\bibitem{zhuge2021salient}
Mingchen Zhuge, Deng-Ping Fan, Nian Liu, Dingwen Zhang, Dong Xu, and Ling Shao.
\newblock Salient object detection via integrity learning.
\newblock {\em IEEE TPAMI}, 2022.

\bibitem{zou2023generalized}
Xueyan Zou, Zi-Yi Dou, Jianwei Yang, Zhe Gan, Linjie Li, Chunyuan Li, Xiyang Dai, Harkirat Behl, Jianfeng Wang, Lu Yuan, et~al.
\newblock Generalized decoding for pixel, image, and language.
\newblock In {\em CVPR}, pages 15116--15127, 2023.

\end{thebibliography}
}
\clearpage

\end{document}